\documentclass[10pt,twocolumn,letterpaper]{article}

\usepackage{iccv}
\usepackage{times}
\usepackage{epsfig}
\usepackage{graphicx}
\usepackage{amsmath}
\usepackage{amssymb}
\usepackage{tabularx}
\usepackage{rotating} 
\usepackage{tablefootnote}
\usepackage{subfigure}
\usepackage{overpic}
\usepackage{capt-of}
\usepackage{colortbl}
\usepackage{xcolor}

\usepackage[breaklinks=true,bookmarks=false]{hyperref}

\newcolumntype{C}[1]{>{\centering\arraybackslash}p{#1}}
\newcommand{\uline}{\underline}

\iccvfinalcopy 

\def\iccvPaperID{744} 

\ificcvfinal\fi
\begin{document}
\title{Flow Fields: Dense Correspondence Fields for Highly Accurate Large Displacement Optical Flow Estimation}

\author{~~~Christian Bailer$^1$~~~~~~~~~~~~~~~~~~~~~~Bertram Taetz$^{1,2}$~~~~~~~~~~~~~~~~~~~Didier Stricker$^{1,2}$ \\
{\tt\small Christian.Bailer@dfki.de}~~~~~~~{\tt\small Bertram.Taetz@dfki.de}~~~~~~~{\tt\small Didier.Stricker@dfki.de}\\
$^1$German Research Center for Artificial Intelligence (DFKI), $^2$University of Kaiserslautern}
\maketitle
\begin{abstract}
Modern large displacement optical flow algorithms usually use an
initialization by either sparse descriptor matching techniques or dense approximate nearest neighbor fields. 
While the latter have the advantage of being dense, they have the major disadvantage of being very outlier prone as they are not designed 
to find the optical flow, but the visually most similar correspondence.
In this paper we present a dense correspondence field approach that is much less outlier prone
and thus much better suited for optical flow estimation than approximate nearest neighbor fields.
Our approach is conceptually novel as it does not require explicit regularization, smoothing (like median filtering) or a new data term,
but solely our novel purely data based search strategy that finds most inliers (even for small objects), while it effectively avoids finding outliers. 
Moreover, we present novel enhancements for outlier filtering. 
We show that our approach is better suited for large displacement optical flow estimation than 
state-of-the-art descriptor matching techniques. We do so by initializing EpicFlow (so far the best method on MPI-Sintel)
with our Flow Fields instead of their originally used state-of-the-art descriptor matching technique.
We significantly outperform the original EpicFlow on MPI-Sintel, KITTI and Middlebury.

\end{abstract}
\section{Introduction}
Finding the correct dense optical flow between images or video frames is a challenging problem.
While the visual similarity between two image regions is the most important clue for finding the optical flow, it
is often unreliable due to illumination changes, deformations, repetitive patterns, low texture, occlusions or blur.
Hence, basically all dense optical flow methods add prior knowledge about the properties of the flow, like local smoothness
assumptions~\cite{horn1981determining}, structure and motion adaptive assumptions~\cite{Wedel2009}, 
the assumption that motion discontinuities are more likely at image edges~\cite{revaud:hal-01097477}, 
or the assumption that the optical flow can be approximated by a few motion patterns~\cite{chen2013large}.
The most popular of these assumptions is the local smoothness assumption. It is usually incorporated into a joint energy based 
regularization that rates data consistency together with the smoothness in a variational setting of the flow~\cite{horn1981determining}.
One major drawback of this setting is that fast minimization techniques usually rely on local linearization of the data term and thus
can adapt the motion field only very locally. Hence, these methods have to use image pyramids 
to deal with fast motions (large displacements)~\cite{brox2004high}. 
In practice, this fails in cases where the determined motion on a lower scale is not very close to the correct motion of a higher scale.

\begin{figure}[t]
\subfigure[ANNF~\cite{he2012computing}]{\includegraphics[width=0.495\linewidth]{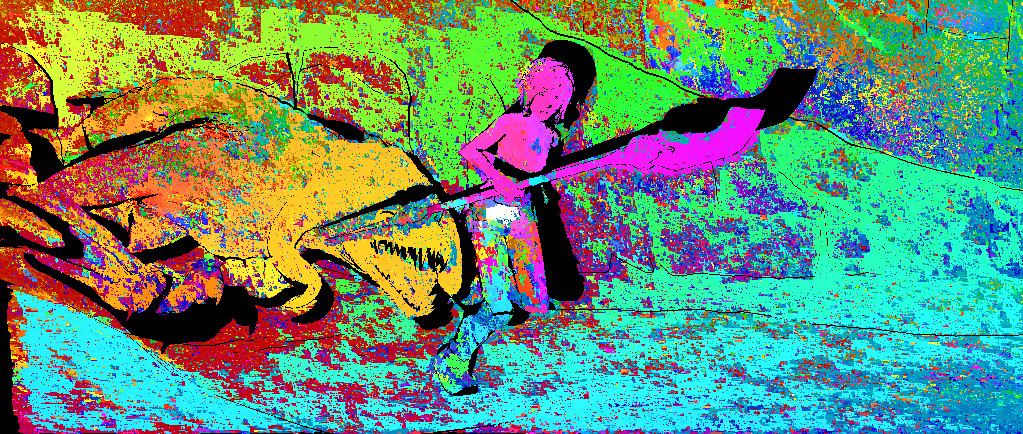}}
\subfigure[Our Flow Field]{\includegraphics[width=0.495\linewidth]{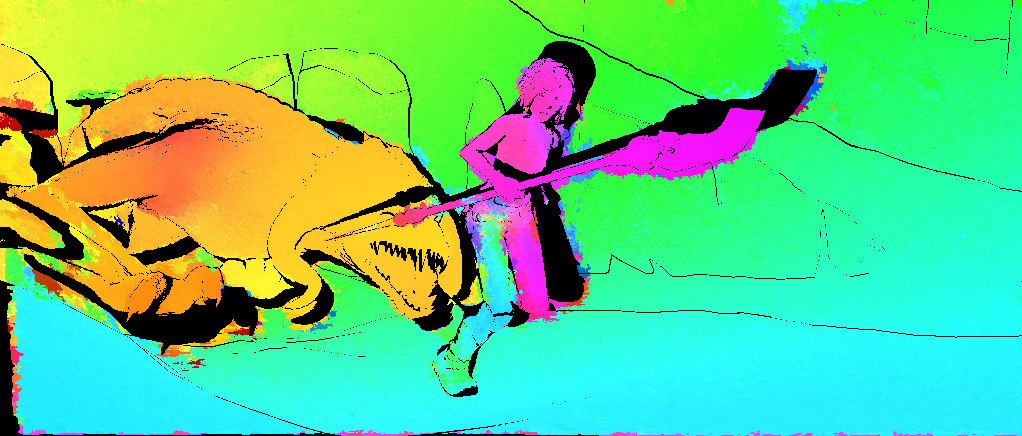}} 
\subfigure[Our outlier filtered Flow Field]{\includegraphics[width=0.495\linewidth]{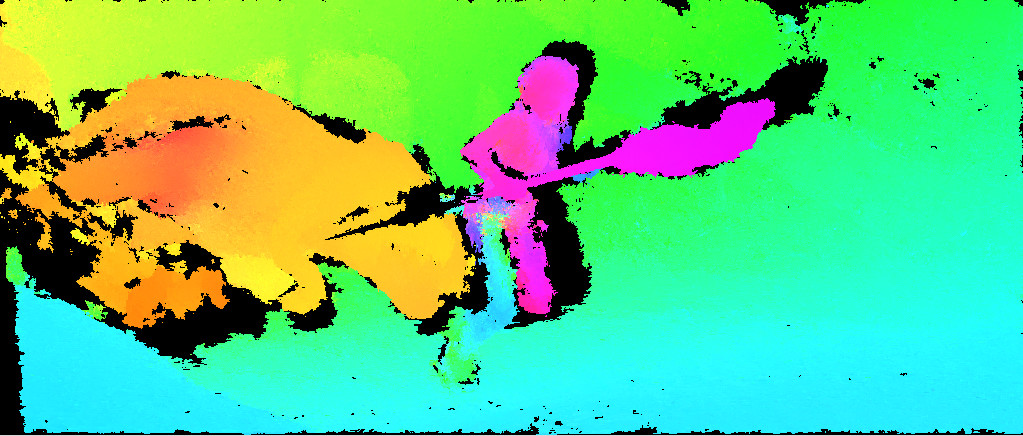}}
\subfigure[Ground truth]{\includegraphics[width=0.495\linewidth]{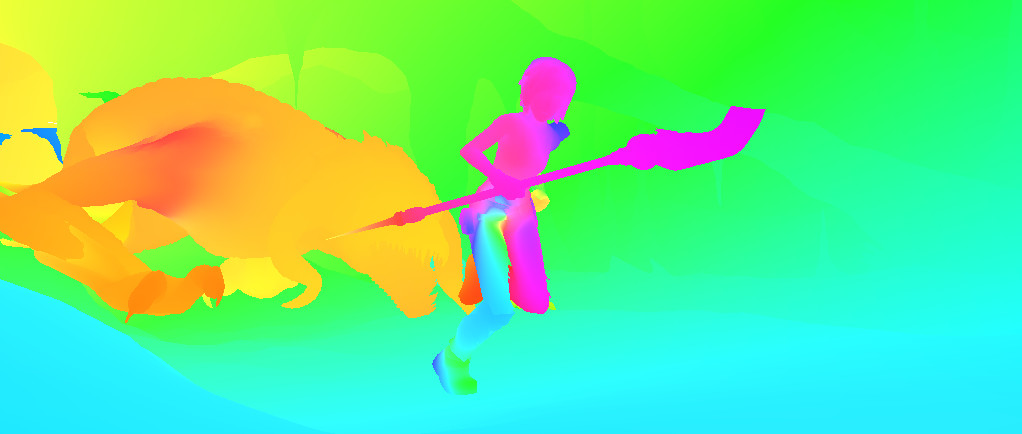}}
 \caption{  Comparison of state-of-the-art approximate nearest neighbor fields (a) and Flow Fields (b) with the same data term. 
 a) and b) are shown with ground truth occlusion map (black pixels).
 c) is after outlier filtering, occluded regions are successfully filtered.
 It can be used as initialization for an optical flow method. }
\label{flowfields}
\end{figure}
In contrast, for purely data based techniques like approximate nearest neighbor fields~\cite{he2012computing} (ANNF) 
and sparse descriptor matches~\cite{weinzaepfel:hal-00873592} 
there are fast approaches that can efficiently perform a global search for the best match on the full image resolution.
However, as there is no regularization, (approximate) nearest neighbor fields (NNF) usually contain many outliers that are difficult to identify. 
Furthermore, even if outliers can be identified they leave gaps in the motion field that must be filled.
Sparse descriptor matches usually contain fewer outliers as matches are only determined for carefully selected points with high confidence. 
However, due to their sparsity the gaps between matches are usually even larger than in outlier filtered ANNF. 
Gaps can be problematic, since a motion for which no match is found cannot be considered. Despite these difficulties, 
ANNF and sparse descriptor matches gained a lot of popularity in the last years as initial step of large displacement optical flow algorithms.
Nowadays, nearly all top-performing methods on challenging datasets like MPI-Sintel~\cite{butler2012naturalistic} rely on such techniques.
However, while there are descriptor matching approaches like Deep Matching~\cite{weinzaepfel:hal-00873592} that are tailored for optical flow, 
dense initialization is usually simply based on ANNF -- which is suboptimal.
The intention behind ANNF is to find the visually closest match (NNF), which is often not identical with the optical flow. 
An important difference is that NNF are known to be very noisy regarding the offset of neighboring pixels, while optical flow is usually locally smooth
and occasionally abrupt (see Figure \ref{flowfields}).

In this paper we show that it is possible to utilize this fact, to create dense correspondence fields that 
contain significantly fewer outliers than ANNF regarding optical flow estimation -- not because of explicit regularization, smoothing (like median filtering) or a different data term, but  
solely because of our novel search strategy that finds most inliers while it effectively avoids finding outliers.
We call them \textit{Flow Fields} as they are  tailored for optical flow estimation, while they are at the same time dense and purely data term based like ANNF.
Flow Fields are conceptually novel as we avoid building on the popular, but for optical flow estimation inappropriate, (A)NNF concept.
Our contributions are:
\begin{itemize}\itemsep2pt
\item A novel hierarchical correspondence field search strategy that features powerful non-locality in the image space (see Figure~\ref{propimg} a)), 
but locality in the flow space (for smoothness) and can utilize hierarchy levels (scales) as effective outlier sieves.
It allows to obtain better results with hierarchies/scales than without, even for tiny objects and other details. 
\item 
We extend the common forward backward consistency check by a novel two way consistency check as well as region and density based outlier filtering. 
\item We show the effectiveness of our approach by clearly outperforming ANNF and by obtaining the best result on  MPI-Sintel~\cite{butler2012naturalistic} and 
the second best on KITTI~\cite{geiger2013vision}. 
\end{itemize}

\section{Related Work \label{rel}}
Dense optical flow research started more than 30 years ago with the work of Horn and Schunck~\cite{horn1981determining}. 
We refer to publications like~\cite{baker2011database, sun2010secrets, vogel2013evaluation} for a detailed overview of optical flow methods 
and the general principles behind it. 

One of the first works that integrated sparse descriptor matching for improved large displacement performance was Brox and Malik~\cite{brox2011large}.
Since then, several works followed the idea of using sparse descriptors~\cite{xu2012motion,weinzaepfel:hal-00873592,kennedy2015optical,timofte2015sparse,revaud:hal-01097477}.
Only few works used dense ANNF instead~\cite{jith2014optical,chen2013large}. 
Chen et al.~\cite{chen2013large} showed  that remarkable results can be achieved on the  Middlebury evaluation portal by extracting
the dominant motion patterns from ANNF. Revaud et al.~\cite{revaud:hal-01097477} compared ANNF to
Deep Matching~\cite{weinzaepfel:hal-00873592} for the initialization of their approach. They found that Deep Matching clearly outperforms 
ANNF. We will use their approach for optical flow estimation and show that this is not the case for our Flow Fields. 
 
An important milestone regarding fast ANNF estimation was Patchmatch~\cite{barnes2010generalized}.
Nowadays, there are even faster ANNF approaches~\cite{korman2011coherency,he2012computing}. 
There are also approaches that try to obtain correspondence fields tailored to optical flow. 
Lu et al.~\cite{lu2013patch} used superpixels to gain edge aware correspondence fields. 
Bao et al.~\cite{bao2014fast} used an edge aware bilateral data term instead. 
While the edge aware data term helps them to obtain good results -- especially at motion boundaries,
their approach is still based on the ANNF strategy to determine correspondences, although it is unfavorable for optical flow.
HaCohen et al.~\cite{hacohen2011non} presented a hierarchical correspondence field approach for image enhancement. 
While it does well in removing outliers, it also removes inliers that are not supported by a big neighborhood (in each scale). 
Such inliers are especially important for optical flow as they cannot be determined by the classical coarse to fine strategy. 
Our approach cannot only preserve such isolated inliers, but can also spread them if needed (Figure~\ref{propimg} a)).
 
A technique that shares the idea of preferring locality (to avoid outliers) with our approach is region growing in 3D reconstruction~\cite{goesele2007multi,furukawa2010accurate}. 
It is usually computationally expensive. 
A faster GPU parallelizable alternative based on PatchMatch~\cite{barnes2010generalized} was presented in our previous work~\cite{bailer2012scale}. 
It shares some ideas with our basic approach in Section~\ref{bas}, but was not
designed for optical flow estimation and lacks many important aspects of our approach in this paper. 

\section{Our Approach \label{our} } 
In this section we detail our Flow Field approach, our extended outlier filter and the data terms used in the tests of our paper. 
Flow Fields are described in two steps. First we describe a basic (non-hierarchical) Flow Field approach. Afterwards, we build
our full (hierarchical) Flow Field approach on top of it.
Given two images $I_1, I_2 \subset \mathbb{R}^2$ we use the following notation:
$P_r(p_i)$ is an image patch with patch radius $r$ centered at a pixel position
$p_i = (x,y)_i \in I_i\,\, i=1,2$. The total size of
our rectangular patch is $(2r+1) \times (2r+1)$ pixels.
Our goal is to determine the optical flow field of $I_1$ with respect to $I_2$
i.e. the displacement field for all pixels $p_1 \in I_1$, denoted by $F(p_1) =
M(p_1)-p_1 \in \mathbb{R}^2$ for each pixel $p_1$.
$M(p_1)$ is the corresponding matching position $p_2 \in I_2$ for a position
$p_1 \in I_1$. All parameters mentioned below are assigned in Section~\ref{eva}.
\subsection{Basic Flow Fields \label{bas}}
The first step of our basic approach is similar to the kd-tree based initialization step of the ANNF approach of He and Sun~\cite{he2012computing}.
We do not use any other step of~\cite{he2012computing} 
as we have found them to be harmful for optical flow estimation, since they introduce \textit{resistant outliers}  
whose matching errors are below those of the ground truth.
Once introduced, a purely data based approach without regularization cannot remove them anymore.
The secret is to avoid finding them.\footnote{ ANNF try to reproduce the NNF that contains all resistant outliers.}

\begin{figure}[t]
\center
 ~
\raisebox{0.85cm}{a)}~~\includegraphics[width=0.23\linewidth]{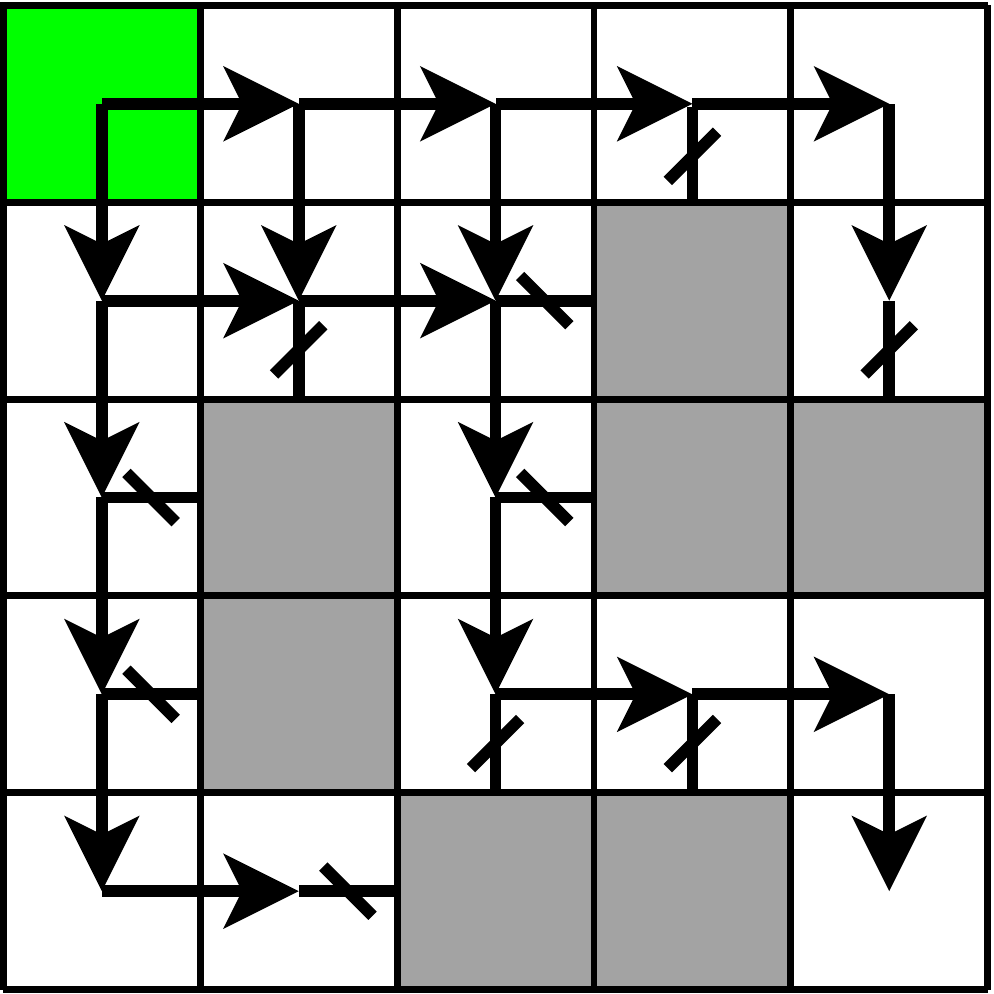}
~~~
 \raisebox{0.85cm}{b)}~~\includegraphics[width=0.23\linewidth]{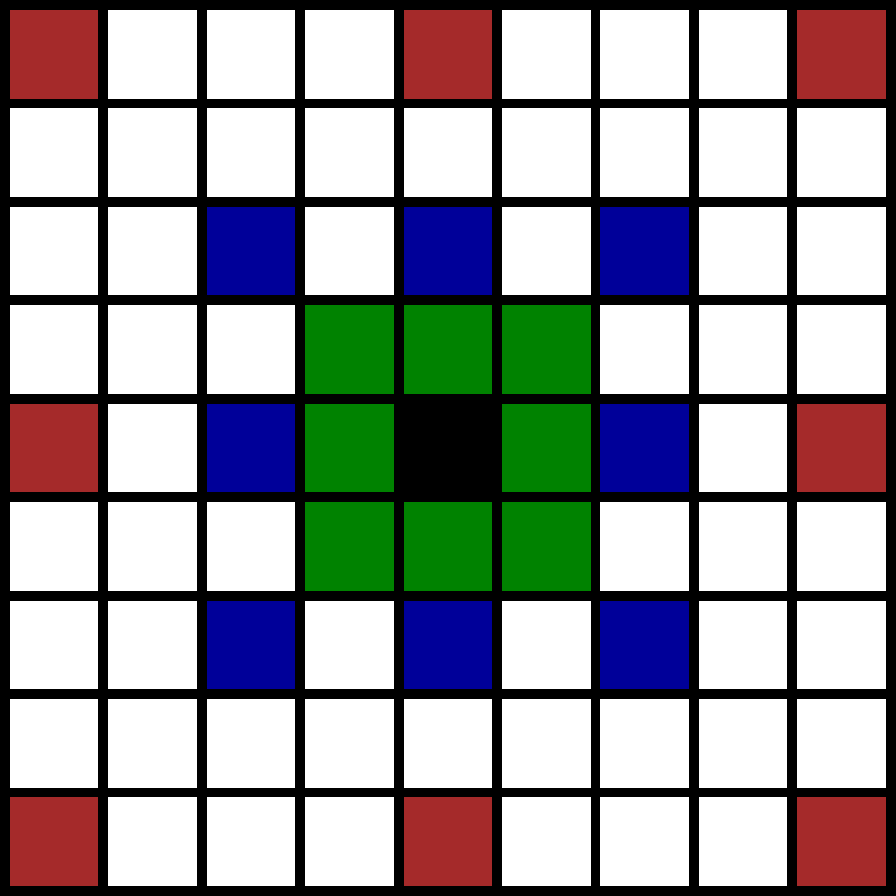}
~~~
 \raisebox{0.85cm}{c)}~~\includegraphics[width=0.25\linewidth]{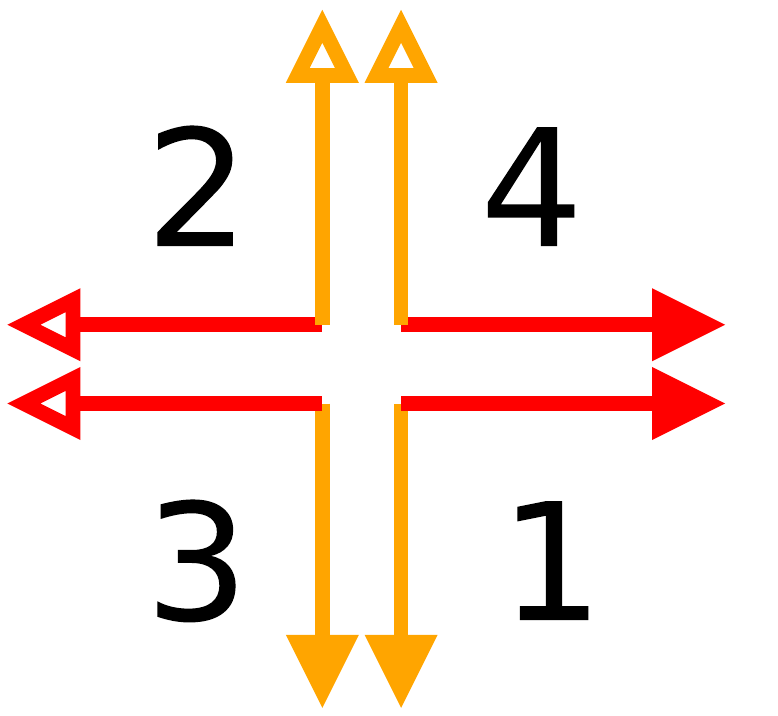}
\caption{a) Example for the ability of propagation to propagate into different directions within a 90 degree angle. Gray pixels
reject the flow of the green seed pixel. In practice each pixel is a seed. b) Pixel positions of $P_1$ (green), $P^2_1$ (blue) and $P^4_1$ (red).
The central pixel is in black. c) Our propagation directions. }
\label{propagation}
\end{figure}

Our approach, outlined in Figure~\ref{pipeline}, works as follows:
\begin{figure}[t]
  \includegraphics[width=1.0\linewidth]{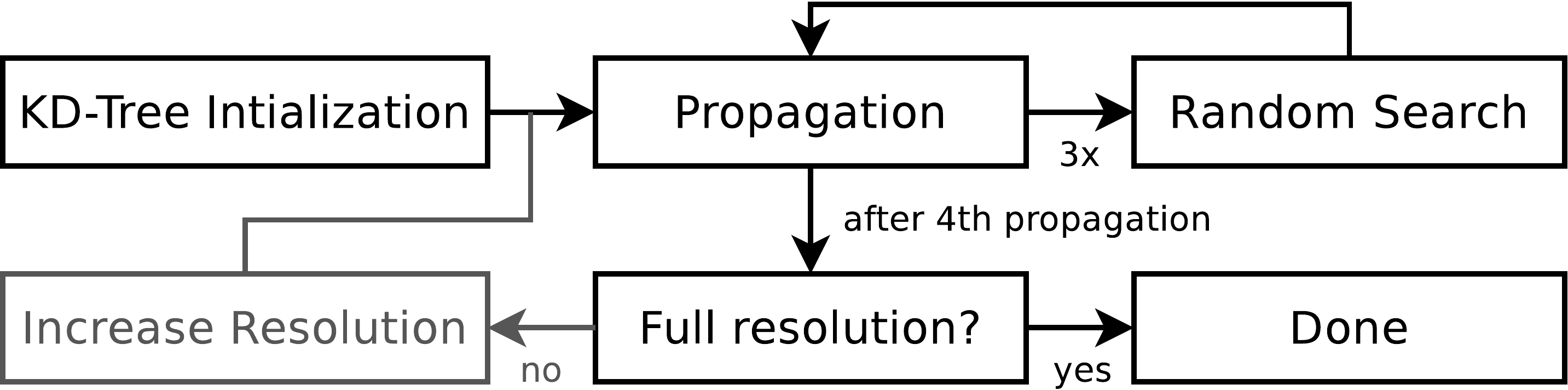}
   \caption{The pipeline of our Flow Field approach. For the basic approach we only consider the full resolution.}
\label{pipeline}
\end{figure}
First we calculate the Walsh-Hadamard Transform (WHT)~\cite{hel2005real} for all patches $P_r(p_2)$ centered 
at all pixel positions $p_2$ in image $I_2$ similar to~\cite{he2012computing}.\footnote{
For WHTs patches must be split in the middle. We found that quality does not suffer from
spiting uneven patches ($2r+1$) into sizes $r$ and $r+1$.
} 
In contrast to them we use the first 9 bases for all three color channels in the CIELab color space.
The resulting 27 dimensional vectors for each pixel are then sorted into a kd-tree with leaf size $l$. 
We also split the tree in the dimension of the maximal spread by the median value. 
After building the kd-tree we create WHT vectors for all patches $P_r(p_1)$ at all pixel positions in image $I_1$ as well
and search the corresponding leaf within the kd-tree (where it would belong to if we would add it to the tree). 
All $l$ entries $L$ in the leaf found by the vector of the patch $P_r(p_1)$ are considered as candidates for the initial Flow Field $F(p_1)$.  
To determine which of them is the best we calculate their matching errors $E_d$ with a robust data term $d$ and only keep the
candidate with the lowest matching error in the initial Flow Field, i.e.
\begin{equation}
F(p_1) = arg\,min_{p_2 \in L} (E_d(P_r(p_1),P_r(p_2) )) - p_1
\end{equation}
This is similar to \textit{reranking} in~\cite{he2012computing}. We call points in the initial Flow Field arising directly from the kd-tree \textit{seeds}.
Larger $l$ increase the probability that both correct seeds and resistant outliers are found. 
However, if both are found at a position the resistant outlier prevails. 
Thus, it is advisable to keep $l$ small and to utilize the local smoothness of optical flow to propagate rare correct seeds 
in the initial Flow Field into many surrounding pixels -- outliers usually fail in this regard as their surrounding does not form a smooth surface.
The propagation of our initial flow values works similar to the propagation step in the PatchMatch approach~\cite{barnes2010generalized} i.e. 
flow values are propagated from position $(x,y-1)_1$ and $(x-1,y)_1$ to position $p_1 = (x,y)_1$ as follows:
\begin{multline} \label{e_pro}
F(p_1) = arg\,min_{p_2 \in G_1 } (E_d(P_r(p_1),P_r(p_2) )) - p_1  \\
G_1 = \{F(p_1),F\big((x,y-1)_1\big),F\big((x-1,y)_1\big)\}+p_1~~
\end{multline}
$G_1$ are the considered flows for our first propagation step.  
It is important to process positions $(x,y-1)_1$ and $(x-1,y)_1$ with Equation~\ref{e_pro} before position $(x,y)_1$ is processed. 
This allows the propagation approach to propagate into arbitrary directions within a 90 degree angle (see Figure~\ref{propagation} a)). 
As optical flow varies between neighboring pixels, but propagation can only propagate existing flow values 
our next step is a random search step. Here, we modify the flow of each pixel $p_1$ by a random 
uniformly distributed offset $O_{rnd}$ of at most $R$ pixels. If the matching error $E$ decreases we replace the flow $F$ by the new flow $F + O_{rnd}$. 
$O_{rnd}$ is a subpixel accurate offset which leads to subpixel accurate positions $M(p_1)$. The pixel colors of $M(p_1)$ and $P_r(M(p_1))$ 
are determined by bilinear interpolation. Early subpixel accuracy not only improves accuracy, but also helps to avoid 
outliers as subpixel accurate matches have a smaller matching error. 

In total we perform alternately 4 propagation and 3 random search steps (all with the same $R$) as shown in Figure~\ref{pipeline}. 
While the first propagation step is performed to the right and bottom, the subsequent three propagation steps are 
performed into the directions shown in Figure~\ref{propagation} c). 
Many approaches that perform propagation (e.g.~\cite{he2012computing}) do not consider different propagation directions. 
Even the original PatchMatch approach only considers the first two directions. While these already include all 4 main directions, we have to consider 
that propagation actually can propagate into all directions within a quadrant (see Figure~\ref{propagation} a)) 
and that there are 4 quadrants in the full 360 degree range.  

Extensive propagation with random search is important to distribute rare correct seeds into the whole Flow Field. 
The locality of single propagation steps and random search (with small $R$) effectively prevents the Flow Field from introducing new outliers
not existing in the initial Flow Field.

\subsection{Flow Fields \label{flo}}

\begin{figure}[t]

  \includegraphics[width=1.0\linewidth]{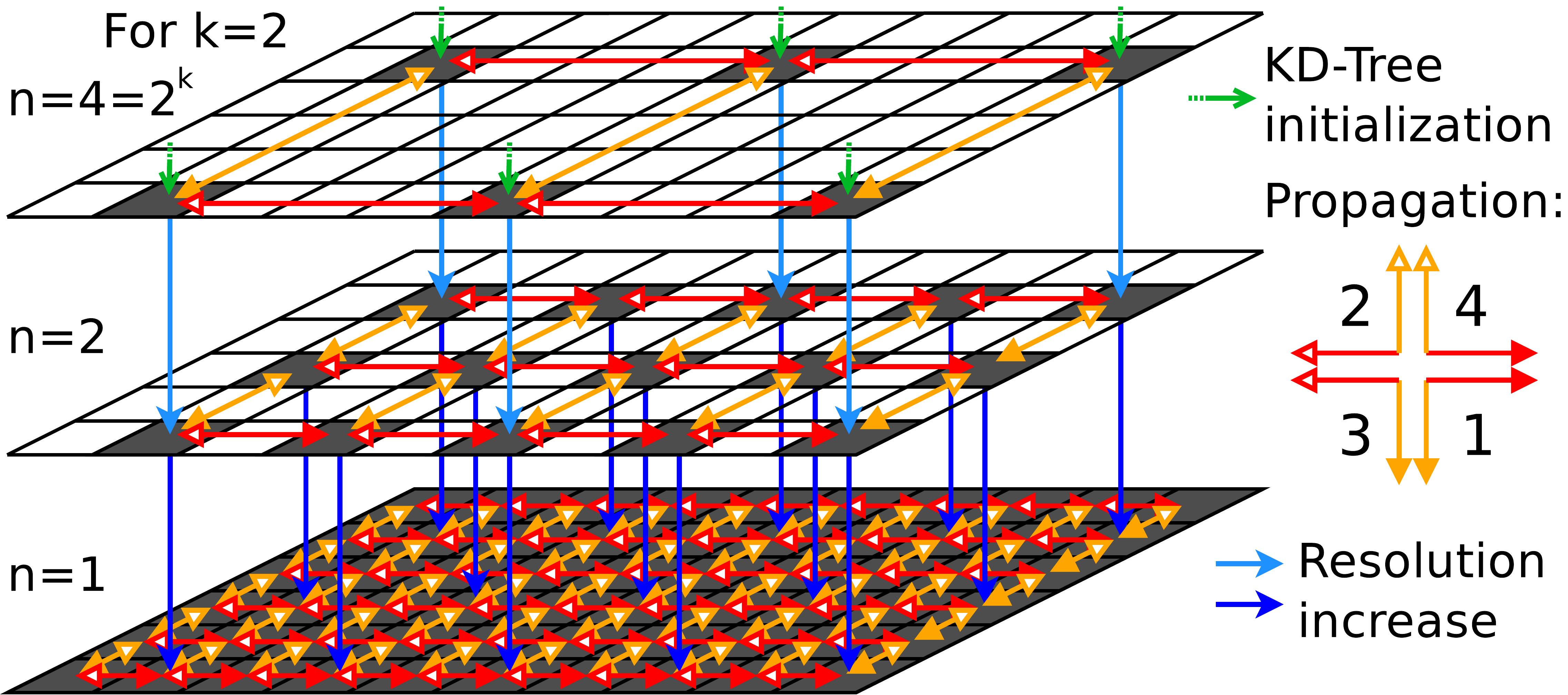}
   \caption{Illustration of our hierarchical Flow Field approach. Flow offsets saved in pixels are propagated in all arrow directions. }
\label{hirarchy}

\end{figure}
Our basic Flow Fields still contain many resistant outliers arising from kd-tree initialization.  
We can further reduce their amount (and the amount of inliers) by not determining an initial flow value for each pixel. 
This helps as inliers usually propagate much further than outliers (optical flow is smooth, outliers are usually not).
However, to cover the larger flow variations between fewer inliers (that are further apart from each other) the random search distance $R$ must be increased,  
which raises the danger of adding close by resistant outliers. A way to avoid this is to increase $r$, as well. 
This helps e.g. in the presence of repetitive patterns or poorly textured regions, but also significantly increases computation time and 
creates new failure cases e.g. close to motion discontinuities and for small objects. 
Furthermore, a larger $r$ (and $R$) leads to less accurate matches.

We found a powerful solution (outlined in Figure~\ref{hirarchy})
that avoids most of the disadvantages of large patches while being even more robust:
First we define that $P^n_r(p_i)$ is a subsampled patch at pixel position $p_i$ with patch radius $r*n$ that 
 consists of only each $n$th pixel within its radius including the center pixel i.e. (see Figure~\ref{propagation} b) for an illustration): 
\begin{equation}
(x^*,y^*) \in P^n_r((x,y)) \Rightarrow \begin{cases} |(x^*-x)|~\text{mod}~n = 0 \\ |(y^*-y)|~\text{mod}~n = 0   \end{cases}
\end{equation} 
The pixel colors for $P^n_r(p_i)$ are not determined from image $I_i$, but from a smoothed version of $I_i$ that we call $I^n_i$.
This is similar to using image pyramids and using $P_r$ on a higher pyramid level. 
The difference is that $I^n_i$ has the full image resolution and that $p_i$ is an actual pixel position on the full resolution, which 
effectively prevents upsampling errors.

As $I^n_i$ only has to be calculated once we can afford to use an expensive low pass filter without noticeable difference in overall processing speed.
In practice, we downsample $I_i$ by a factor of $n$ with area based downsampling, before upsampling it again with Lanczos interpolation~\cite{duchon1979lanczos} 
to obtain $I^n_i$. We always start with $n = 2^k$. Our full Flow Field approach first initializes only each $n$th pixels $p^n_1 =(x_n,y_n)_1$ with
$x_n\mod n = 0$ and $y_n\mod n = 0$ (see Figure \ref{hirarchy}). 
Initialization is performed similar to the basic approach: 
\begin{equation}
 F(p^n_1) = arg\,min_{p_2 \in L} \Big(E_d\big(P^n_r(p^n_1),P^n_r(p_2) \big)\Big) - p^n_1 
\end{equation}
Note that the kd-tree samples $L$ are identical to those of the basic approach. 
We still use non-subsampled patches $P_r(p_i)$ for the WHT vectors for an accurate initialization.

\begin{figure}[t]
 \begin{picture}(0,50)
  \put(0,0){\includegraphics[width=0.495\linewidth]{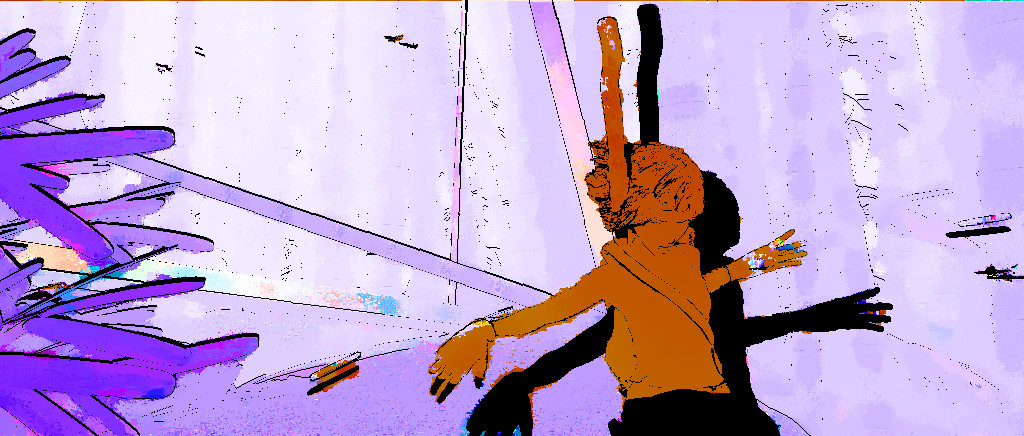}} 
  \put(118,0){\includegraphics[width=0.495\linewidth]{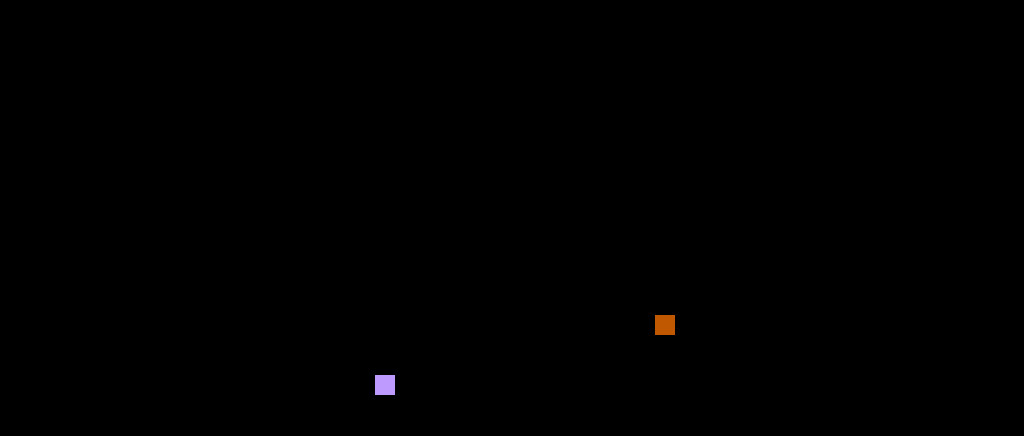}}
   \put(-1,41){\colorbox{white}{\color{black}a)}} 
   \put(117.6,41){\colorbox{white}{\color{black}b)}}
  \end{picture}
 
 \begin{picture}(0,50)
  \put(0,0){\includegraphics[width=0.495\linewidth]{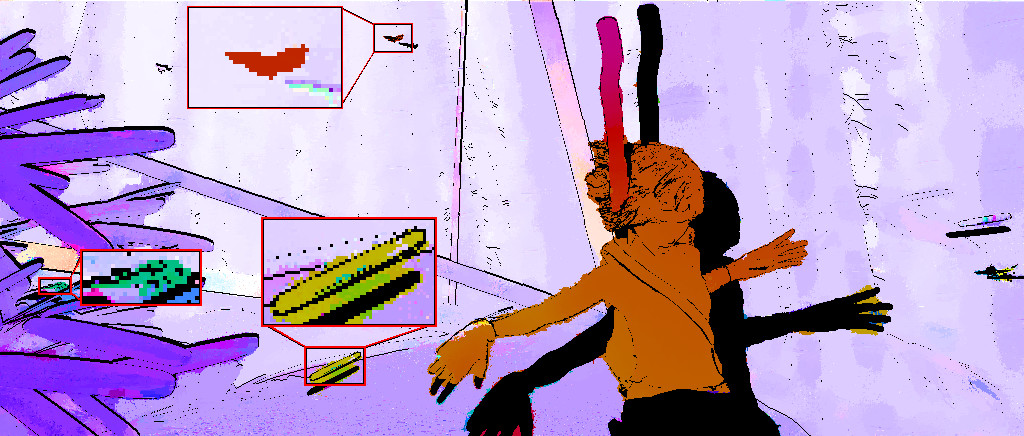}} 
  \put(118,0){\includegraphics[width=0.495\linewidth]{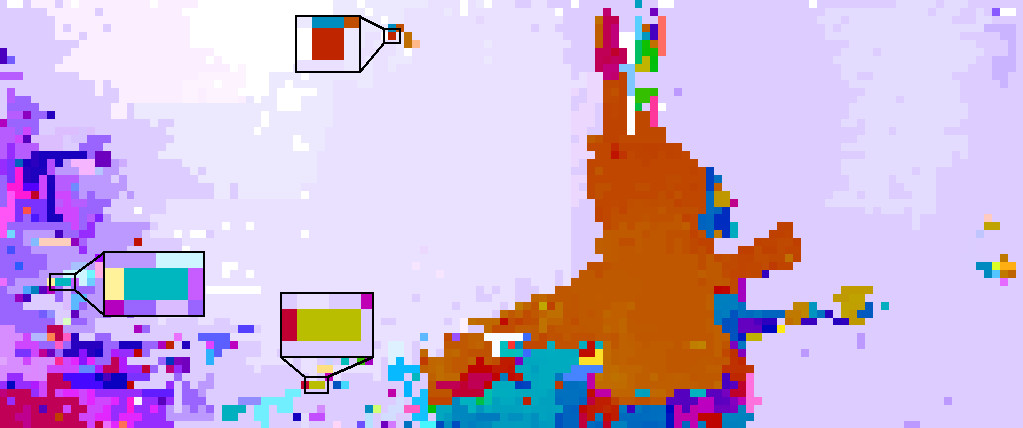}}
   \put(-1,41){\colorbox{white}{\color{black}c)}} 
   \put(117.6,41){\colorbox{white}{\color{black}d)}}
  \end{picture}
  
   \begin{picture}(0,50)
  \put(0,0){\includegraphics[width=0.495\linewidth]{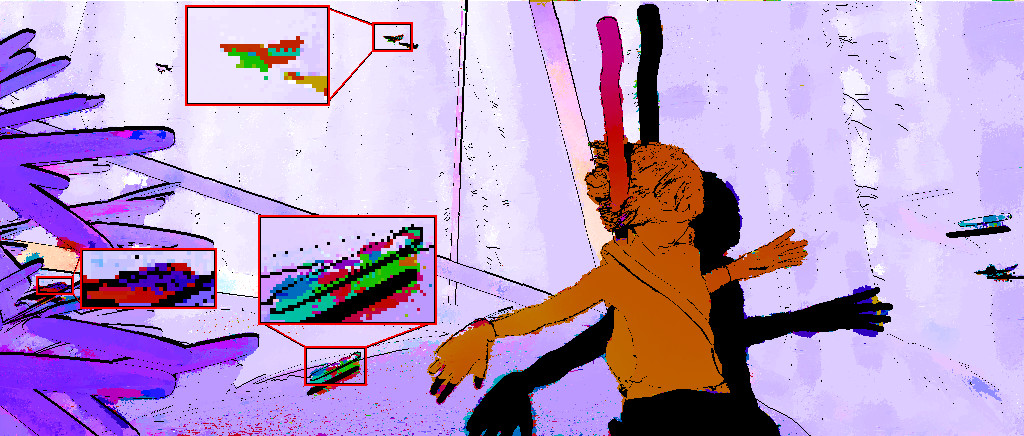}} 
  \put(118,0){\includegraphics[width=0.495\linewidth]{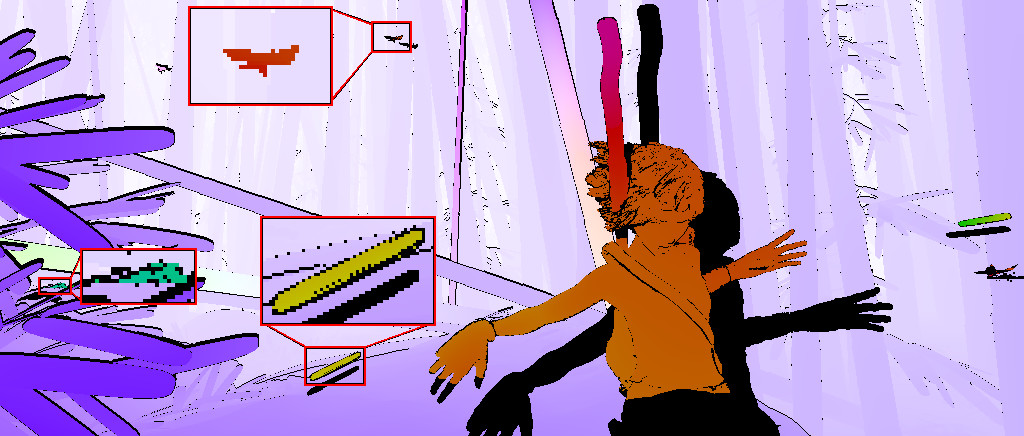}}
   \put(-1,41){\colorbox{white}{\color{black}e)}} 
   \put(117.6,41){\colorbox{white}{\color{black}f)}}
  \end{picture}

  \caption{ \textbf{a)} Flow Field obtained with $k$ $=$ $3$ with b) as  only initialization (black pixels in b) are set to infinity).
   It shows the powerfulness of our hierarchical propagation. 
   \textbf{c)} Like a) but with kd-tree initialization. The 3 marked details are preserved due to their availability
   in the coarse level d). \textbf{e)} like c) but without hierarchies (basic approach). Details are not preserved. 
   \textbf{f)} ground truth. Note: As correspondence estimation is impossible in occluded areas and as orientation we blacked such areas out.}
\label{propimg}
\end{figure}
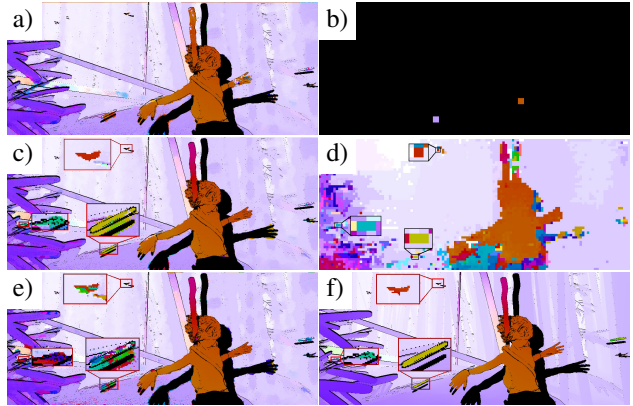
After initialization we perform propagation and random search similar to the basic approach. Except that we only propagate between
points $p^n_1$ i.e.  $(x_n-n,y_n)_1,(x_n,y_n-n)_1 \rightarrow (x_n,y_n)_1$ etc. (see Figure~\ref{hirarchy}) and 
that we use $R_n = R*n$ as maximum random search distance. 
After determining  $F(p^n_1)$ using patches $P^n$, we determine $F(p^m_1), m = 2^{k-1}$ in the same way 
using patches $P^m$. Hereby, the samples $F(p^n_1)$ are used as seeds instead of kd-tree samples.
Positions $p^m_1$ that are not part of $p^n_1$ receive an initial flow value in the first propagation step of the hierarchy level $k-1$. 
This approach is repeated up to the full resolution $F(p^1_1) = F(p_1)$ (see Figure~\ref{pipeline} and \ref{hirarchy}).  

Propagation and random search (with small enough $R$) are usually too local in flow space to introduce new outliers, while 
propagations of lower hierarchy levels are likely to remove most outliers persisting in higher levels,
since resistant outliers are often not resistant on all levels. Thus, hierarchy levels serve as effective outlier sieves (see videos in supplementary material).
Also, matching patches $P^n_r$ is mostly significantly more robust than matching patches $P_{r*n}$ if $r$ is sufficiently large. 
Deformations affect smoothed patches e.g. less, as smoothing allows more matching inaccuracy for a good match.  
Still, we obtain accurate flow values as we are iteratively increasing the resolution.

In contrast to ordinary multi scale approaches, our hierarchical approach is non-local in the image space. 
Figure~\ref{propimg} a) demonstrates how powerful this non-locality is. The Flow Field is only initialized by two flow values with a flow offset of 52 pixels 
to each other (Figure~\ref{propimg} b)). This is more than the random search step of all hierarchy levels together can traverse. 
Thus, the orange flow is a propagation barrier for the violet flow (Like gray pixels in Figure~\ref{propagation} a)). 
Anyhow, our approach manages to distribute the violet flow and similar flows determined by random search throughout the whole image. 
We originally performed the experiment to prove that the flow can be propagated into the arms starting from the body, but 
our approach even can obtain the flow for nearly the whole image with such poor initialization.

Figure~\ref{propimg} c) shows that we can even find tiny objects with our hierarchical approach: 
The 3 marked objects are well persevered in c) due to their availability in the coarse level d).
Remarkably, these objects are only preserved when using hierarchical matching.  
Our basic approach without hierarchies only preserves parts of the upper object (a butterfly) riddled with outliers,
although its seeds are a superset of the seed of the hierarchical approach -- but it fails in avoiding resistant outliers.
Our hierarchical approach preserves tiny objects due to unscaled WHTs (initialization) and since 
the image gradients around tiny objects create local minima in $E_d$, even for huge patches $P^n_r$. This is sufficient
as lower minima (resistant outliers) are successfully avoided by our search strategy.   
Our visual tests showed that our approach with $k=3$ in general preserves tiny objects and other details better
than our basic approach. With too large  $k$ ($>3$) tiny objects are due to lack of seeds not that well preserved.

\subsection{Data Terms \label{dat}}

In our paper we consider two data terms:
\begin{enumerate}\itemsep2pt
\item Census transform~\cite{zabih1994non}. It is computationally cheap, illumination resistant and to some extend edge aware.
We use the sum of census transform errors over all color channels in the CIELab color space for $E_d$.
\item Patch based SIFT flow~\cite{liu2011sift}.
A SIFT flow pixel usually has $S = 3$ channels.
The colors are determined by first calculating the 128 dimensional SIFT vector for each pixel and then reducing it by PCA to $S$ dimensions.
The error between Sift Flow colors is determined by the $L_2$ distance. For the images $I^n_i$ we found it advantageous to smooth the Sift Flow images 
 as described in Section~\ref{flo} and to not use larger SIFT features instead. WHTs are still calculated in the CIELab color space.
\end{enumerate}
\subsection{Outlier Filtering \label{out}}
A common approach of outlier filtering is to perform a forward backward consistency check.
We found that the robustness can be improved by a consistency check between two Flow
Fields with different patch radii, as outliers often diverge into different directions.
Practically, we calculate a backward flow for two patch radii $r$ and $r_2$ and 
delete a pixel if it is not consistent to both backward flows i.e. if 
\begin{equation}
 | F(p_1) + F^b_j(p_1+F(p_1))| < \epsilon, j\in {1,2} 
\end{equation}
is not fulfilled for one of the two backward flows $F^b_j$. 
For a 3 way check an additional forward flow could be added, but for a 2 way check an extra 
backward flow performs better (see supplementary material for an explanation). 
 
After the consistency check many of the remaining outliers form small regions that were originally connected to removed outliers.
Thus, we remove these regions as follows: First, we segment the partly outlier filtered Flow Field into regions. 
Neighboring pixels belong to the same region if the difference between their flow is below 3 pixels.\footnote{ 
Only the flow differences between neighboring pixels count. The flow values of a region can vary by an arbitrary offset.}
Then, we test for regions with less than $s$ pixels if it is possible for that region
to add at least one outlier that was removed by the consistency check with the same rule. 
If this is possible, we found a small region that was originally
connected to an outlier and we remove all points in that region. 

\subsection{Sparsification and Dense Optical Flow\label{spa}}
To fill the gaps created by outlier filtering we use the edge preserving interpolation approach proposed by Revaud et al.~\cite{revaud:hal-01097477} (EpicFlow). 
We found that EpicFlow does not work very well with too dense samples. 
Thus, we select one sample in each 3x3 region in the outlier filtered Flow Field if the region still contains at least $e$ samples.
This is our last consistency check. We found that even after region based filtering most remaining outliers are in sparse regions where 
most flow values were removed. The sample that is selected is the sample for which the sum of both forward backward consistency check errors is the smallest.

\newlength {\alleyLeft }
\setlength{\alleyLeft}{13.2cm}
\newlength {\alleyRight}
\setlength{\alleyRight}{9.0cm}
\newlength {\alleyLeftx }
\setlength{\alleyLeftx}{9.0cm}
\newlength {\alleyRightx}
\setlength{\alleyRightx}{3.5cm}

\renewcommand{\tabcolsep}{0.3pt}
\renewcommand{\arraystretch}{0.2}
 
\begin{table*}
\small
 \centering
 \begin{tabular}{C{0.30cm}ccccC{0.32cm}cc}

 \centering ~~~\begin{rotate}{90}~~~~~Images\end{rotate}~~ &  \includegraphics[width=0.235\linewidth]{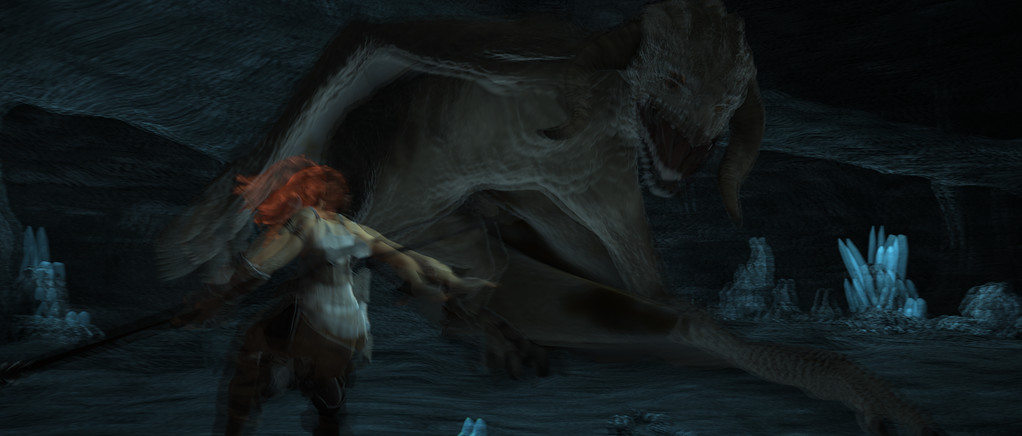} &
 \includegraphics[width=0.235\linewidth]{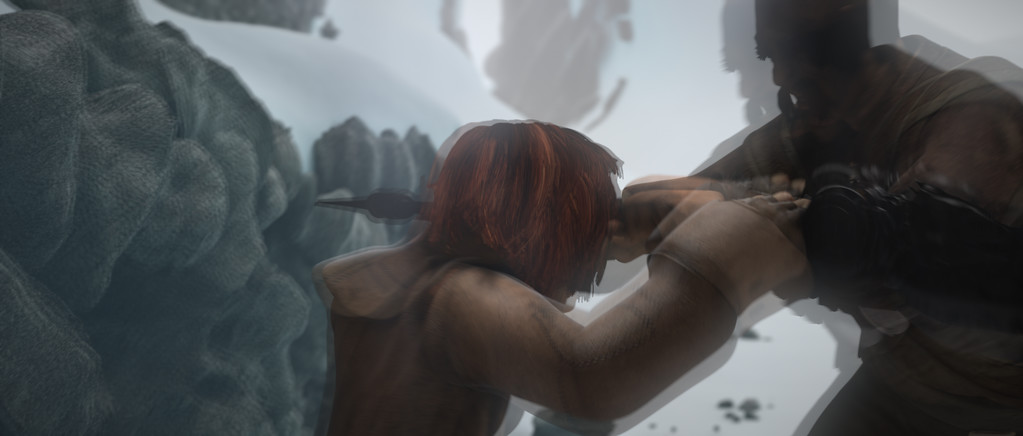}   
 & \includegraphics[height=0.100006\linewidth,trim={ {\the\alleyLeft} 0 {\the\alleyRight} 0},clip]{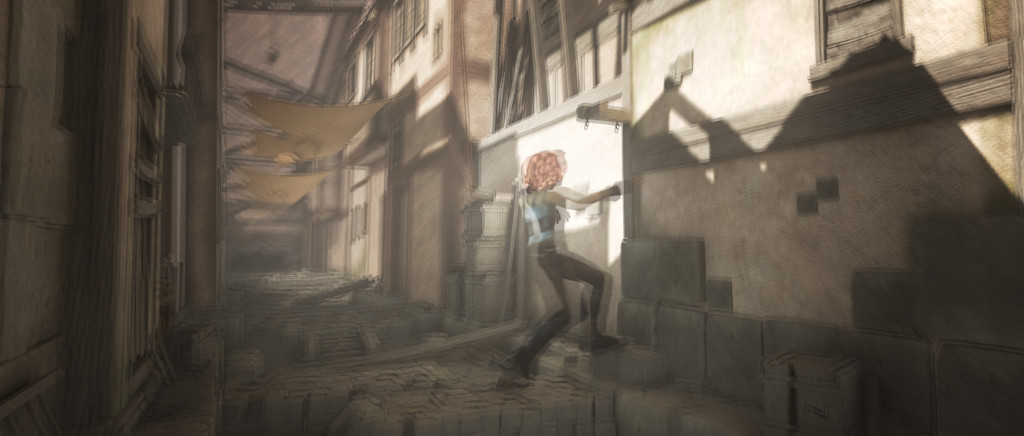} 
 & \includegraphics[height=0.100006\linewidth,trim={ {\the\alleyLeftx} 0 {\the\alleyRightx} 0},clip]{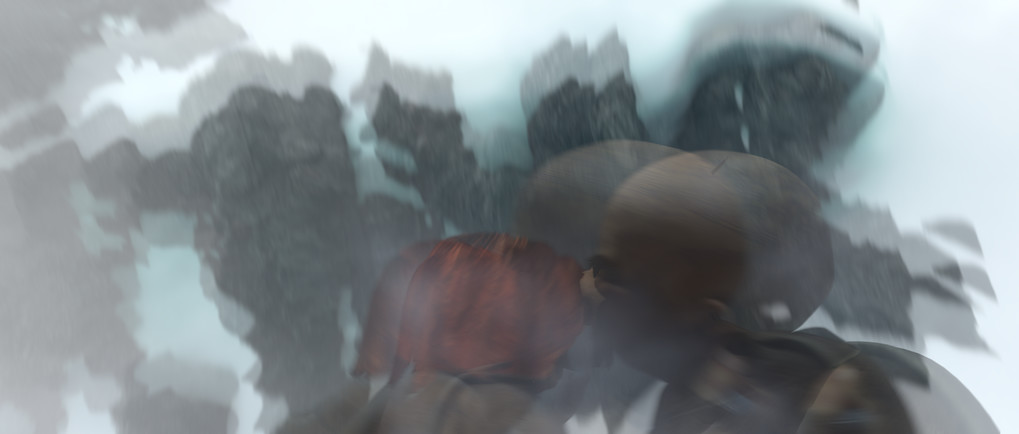}

 &  \centering ~~~~~~\begin{rotate}{90}~\footnotesize ~Defocus blur\end{rotate}~~& 
  \begin{picture}(122.5,0)
    \put(0,0){
   \begin{tabular}[b]{C{2.05cm}C{2.05cm}  }
  \includegraphics[width=\linewidth]{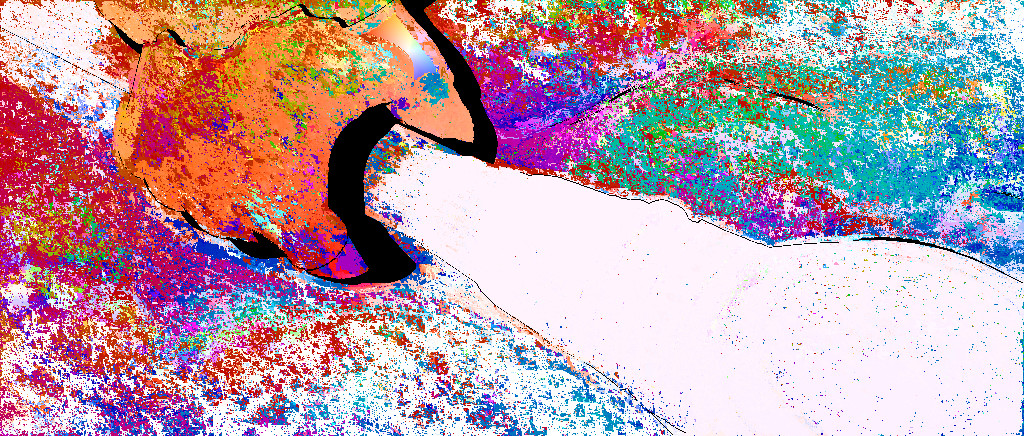} &
 \includegraphics[width=\linewidth]{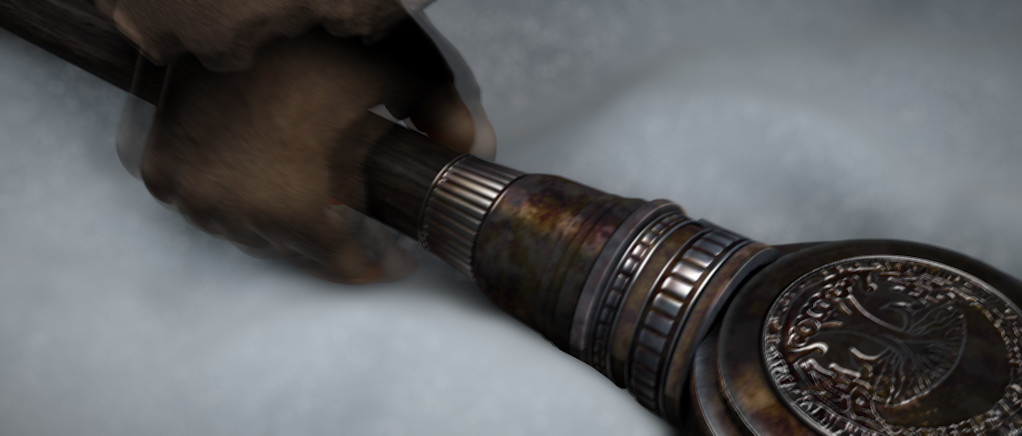} \\
 \includegraphics[width=\linewidth]{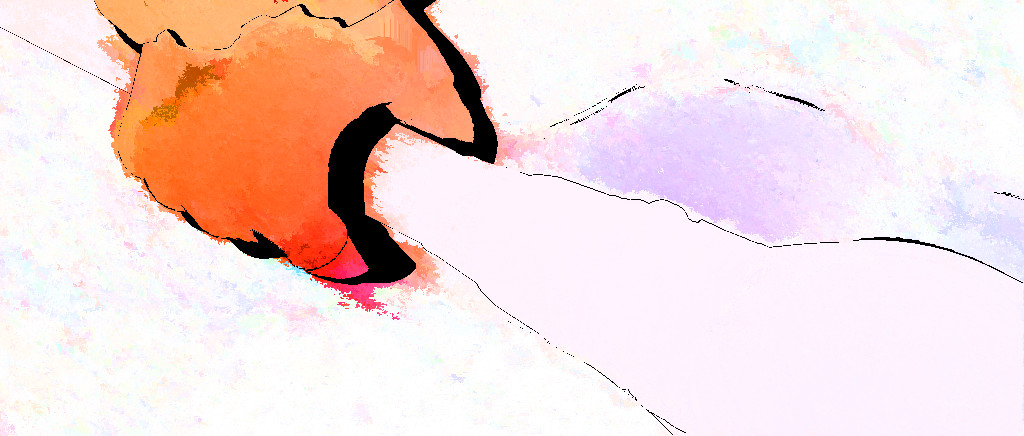} &
 \includegraphics[width=\linewidth]{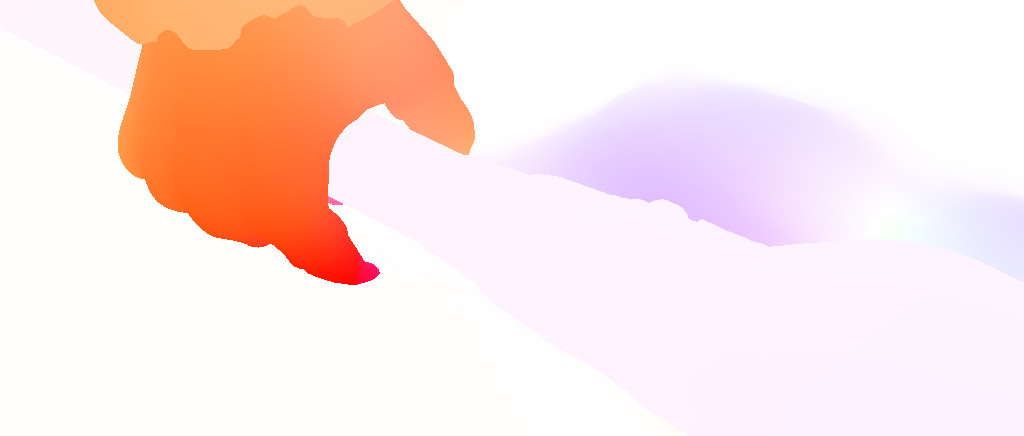} \\
 \end{tabular}}
 \put(0,26.5){\scriptsize ~~~~~~~~~~~ ANNF + OM ~~~~~~~~~~~~~~~~~~~~~\color{white} Images } 
 \put(0,2){\scriptsize ~~~~~~~~~~~~~~~~~ FF + OM ~~~~~~~~~~~ Ground truth } 
  \end{picture}\\
 
 \centering ~~~\begin{rotate}{90}~ANNF+OM\end{rotate}~~ &  \includegraphics[width=0.235\linewidth]{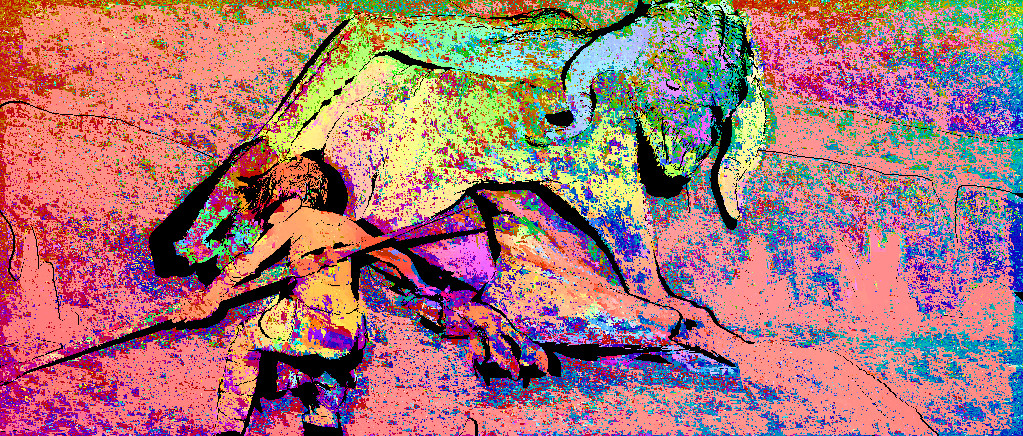} &
 \includegraphics[width=0.235\linewidth]{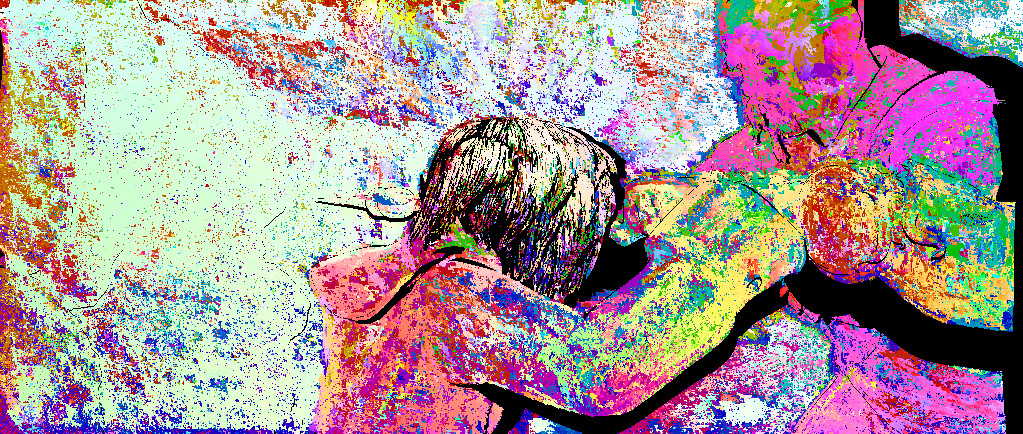}  
 & \includegraphics[height=0.100006\linewidth,trim={ {\the\alleyLeft} 0 {\the\alleyRight} 0},clip]{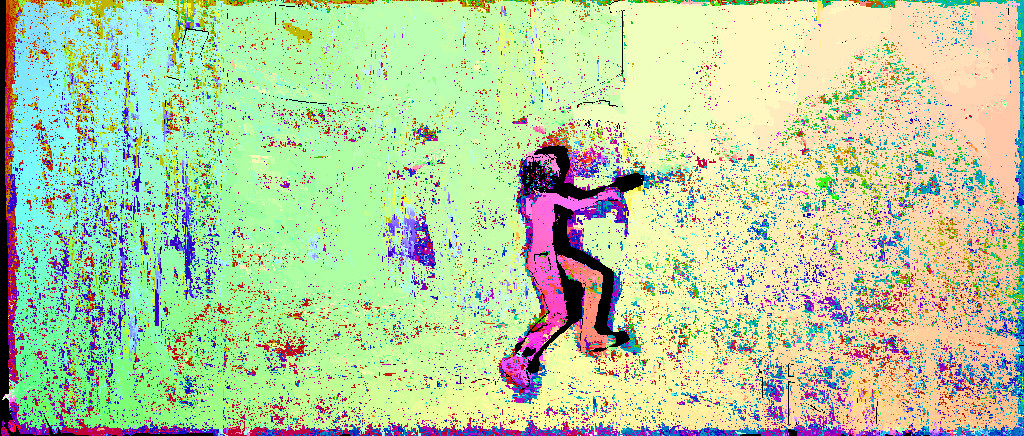} 
 & \includegraphics[height=0.100006\linewidth,trim={ {\the\alleyLeftx} 0 {\the\alleyRightx} 0},clip]{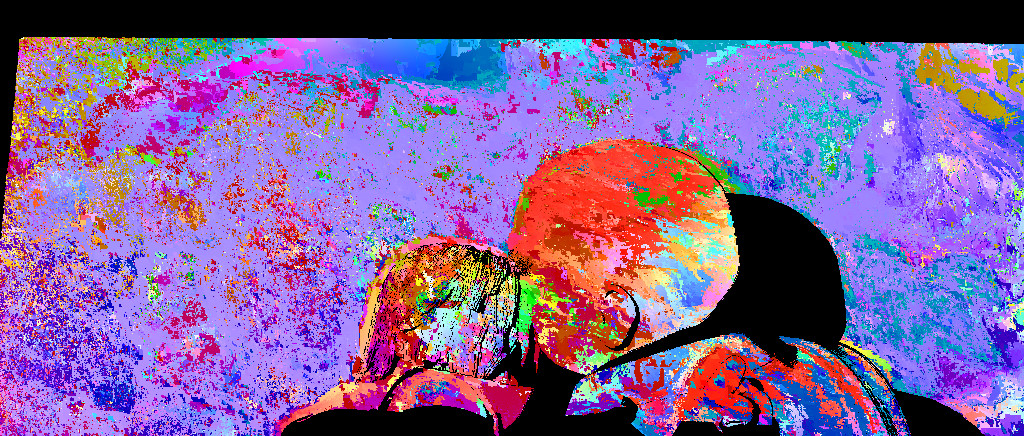}  

 &\centering ~~~~~~\begin{rotate}{90}\textbf{c)}\footnotesize~Motion blur\end{rotate}~~ & 
  \begin{picture}(122.5,0)
    \put(0,0){
 \begin{tabular}[b]{C{2.05cm}C{2.05cm}  }
 \includegraphics[width=\linewidth]{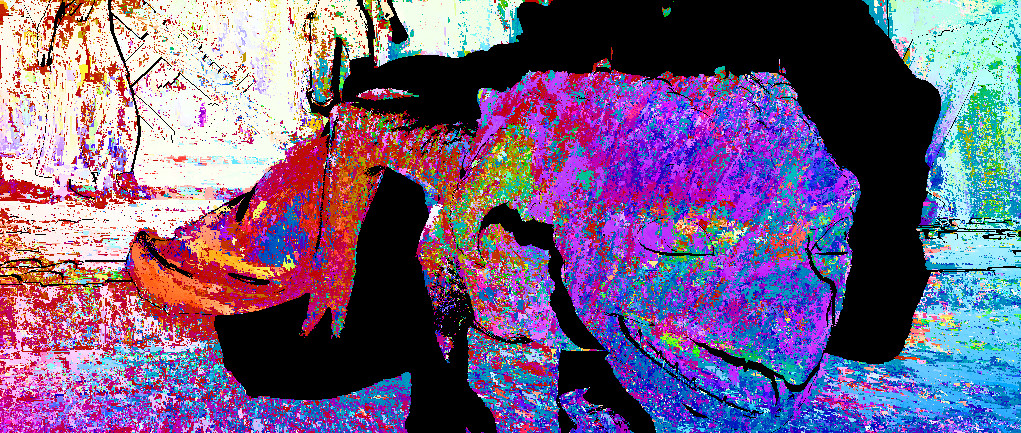} & \includegraphics[width=\linewidth]{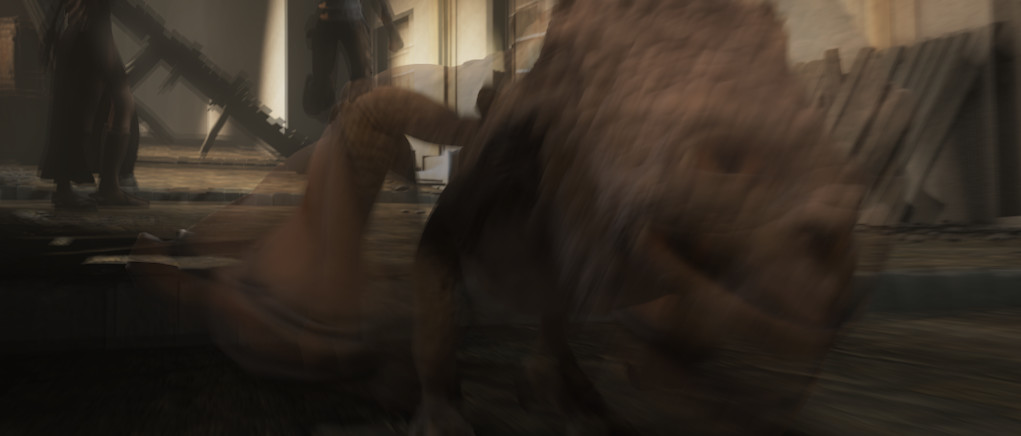}\\
 \includegraphics[width=\linewidth]{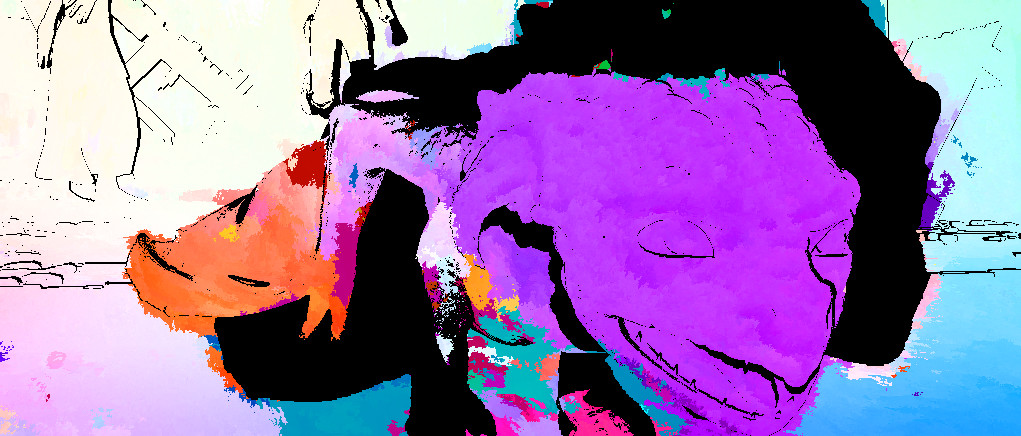} & 
 \includegraphics[width=\linewidth]{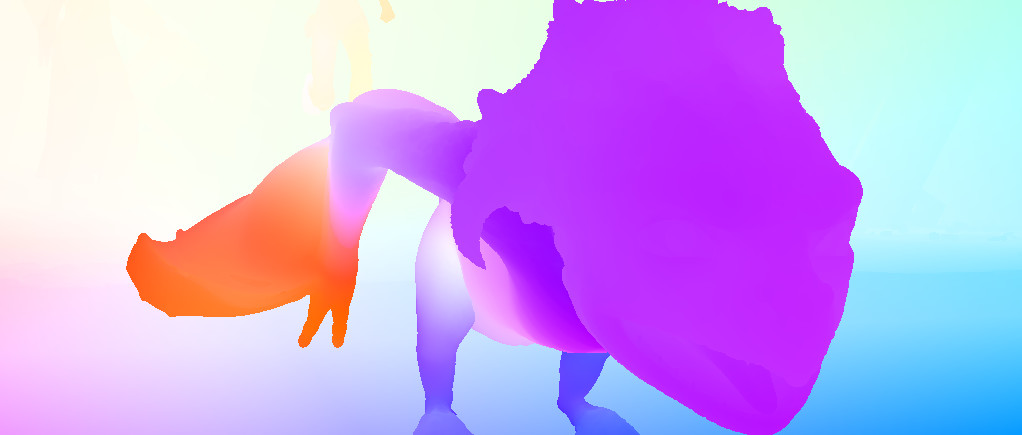} 
 \end{tabular}}
  \put(0,26){\scriptsize~~\color{white}ANNF + OM \color{white} ~~~~~~~~~~ Images } 
 \put(0,0.3){\scriptsize \color{black}~ FF +\color{white} OM \color{black} ~~~~~~~~~~~~~~~~~ Ground truth } 
 \end{picture}\\
  
 \centering ~~~\begin{rotate}{90}~~~~FF+OM\end{rotate}~~ &  \includegraphics[width=0.235\linewidth]{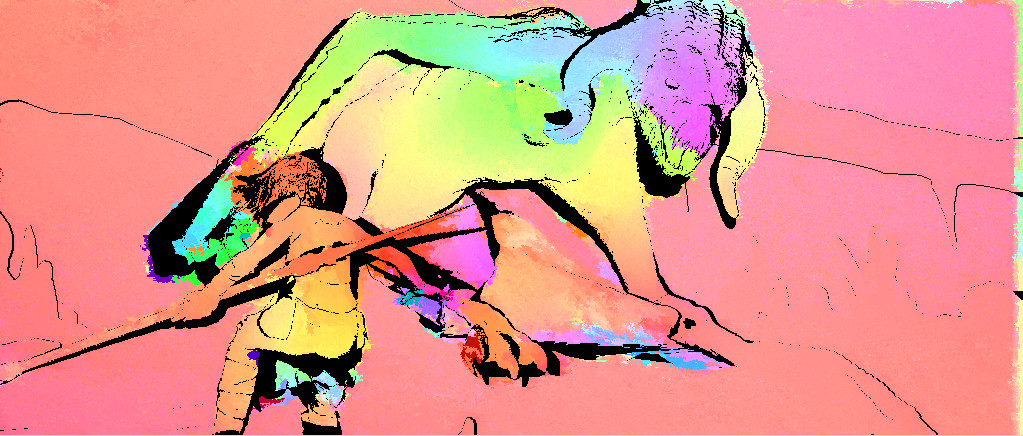} &
 \includegraphics[width=0.235\linewidth]{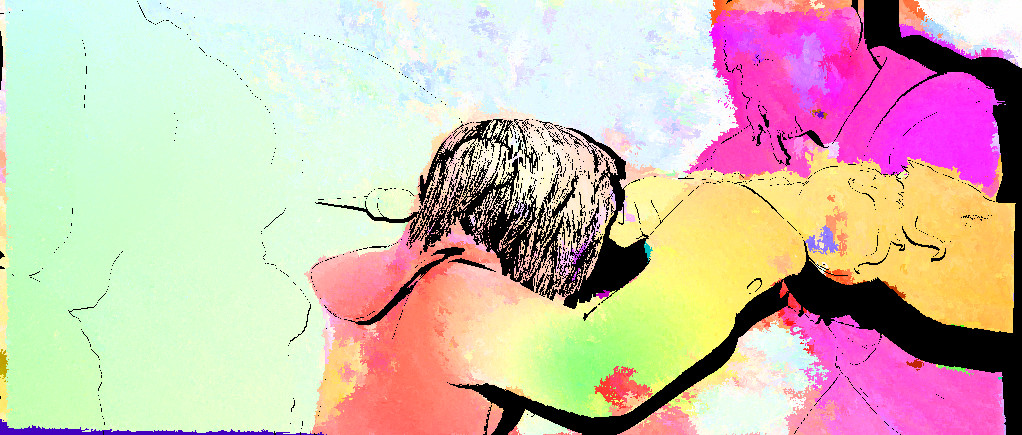} 
 & \includegraphics[height=0.100006\linewidth,trim={ {\the\alleyLeft} 0 {\the\alleyRight} 0},clip]{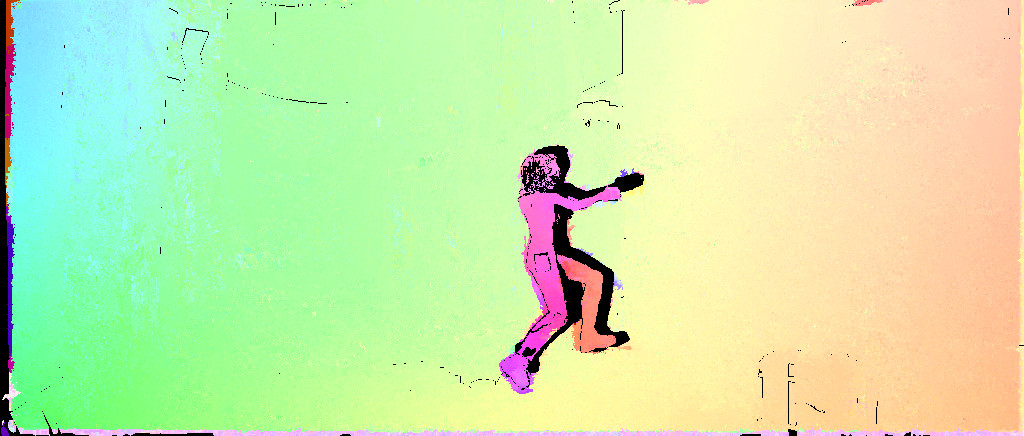}   
 & \includegraphics[height=0.100006\linewidth,trim={ {\the\alleyLeftx} 0 {\the\alleyRightx} 0},clip]{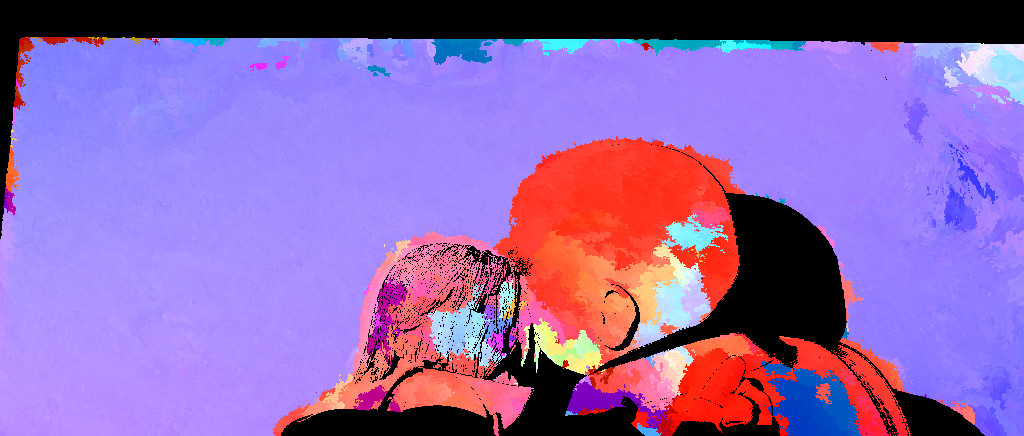} 

 &  \centering ~~~~~\begin{rotate}{90}~~~~~\end{rotate}~~& 
  \begin{picture}(117,0)
 \put(0,0){\includegraphics[width=0.235\linewidth]{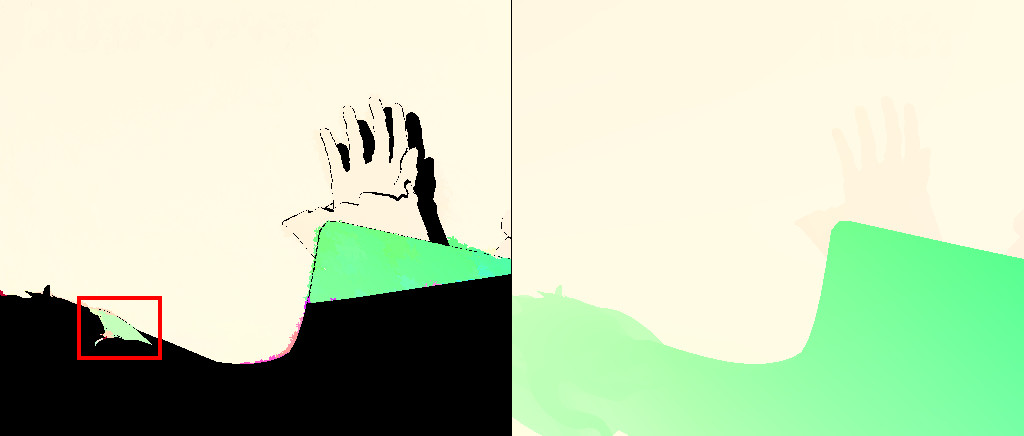}} 
 \put(3,41){ ~~~FF+OM~~~~~~~~~~Ground truth} 
 \end{picture} \\
 
 \centering ~~~\begin{rotate}{90}~~Filtered FF\end{rotate}~~ &  \includegraphics[width=0.235\linewidth]{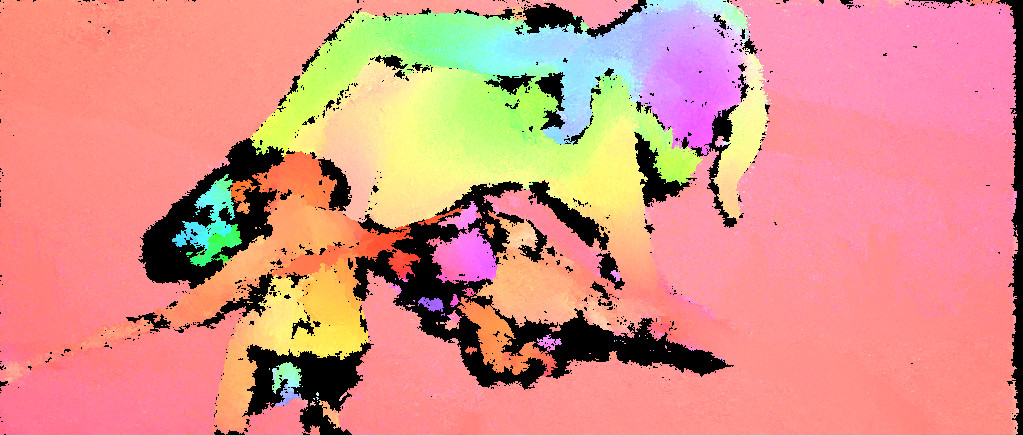} &
 \includegraphics[width=0.235\linewidth]{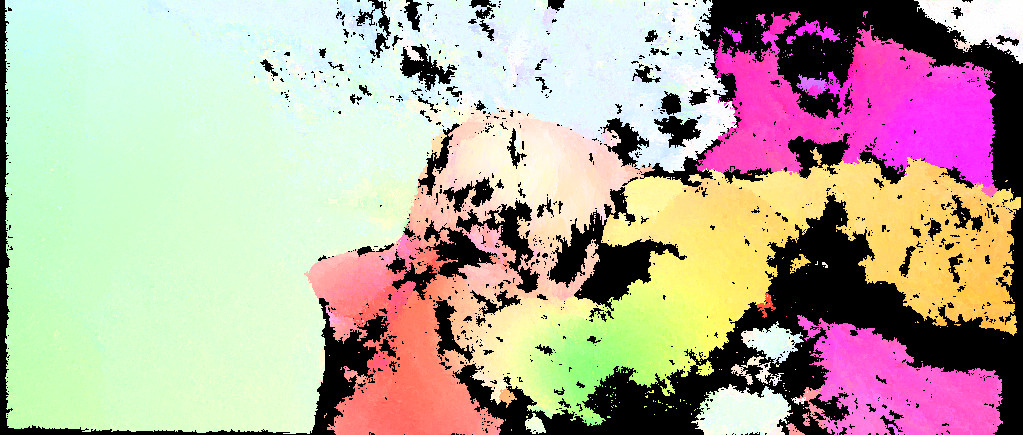} 
 & \includegraphics[height=0.100006\linewidth,trim={ {\the\alleyLeft} 0 {\the\alleyRight} 0},clip]{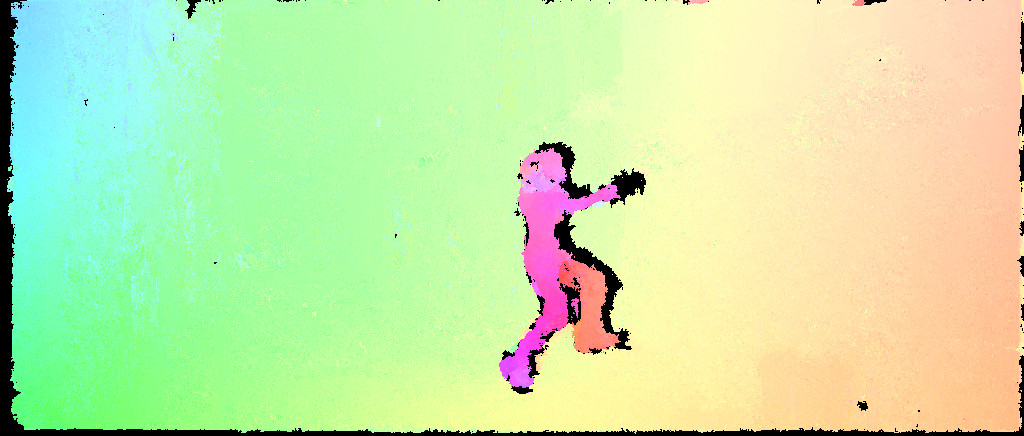}   
 & \includegraphics[height=0.100006\linewidth,trim={ {\the\alleyLeftx} 0 {\the\alleyRightx} 0},clip]{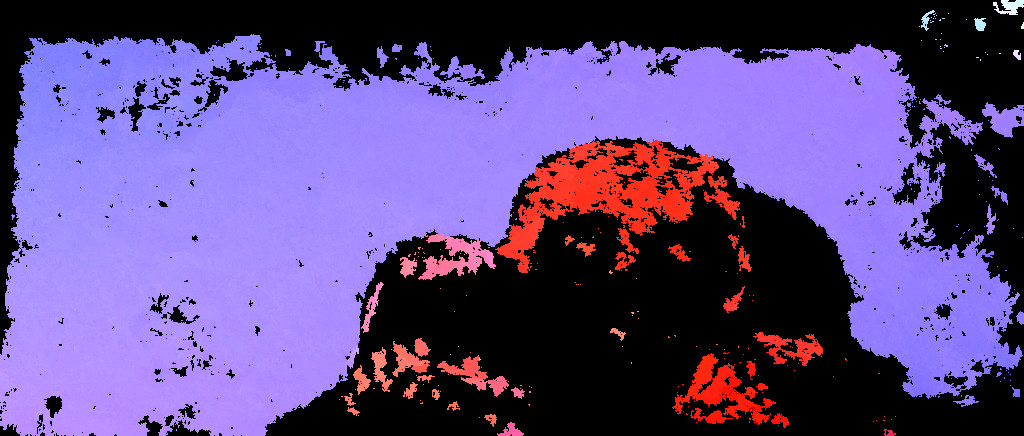} 

 &  \centering ~~~~~~\begin{rotate}{90}\textbf{b)}\footnotesize~~~~Marked detail is filtered \end{rotate}~~&
 \begin{picture}(117,0)
 \put(0,0){\includegraphics[width=0.235\linewidth]{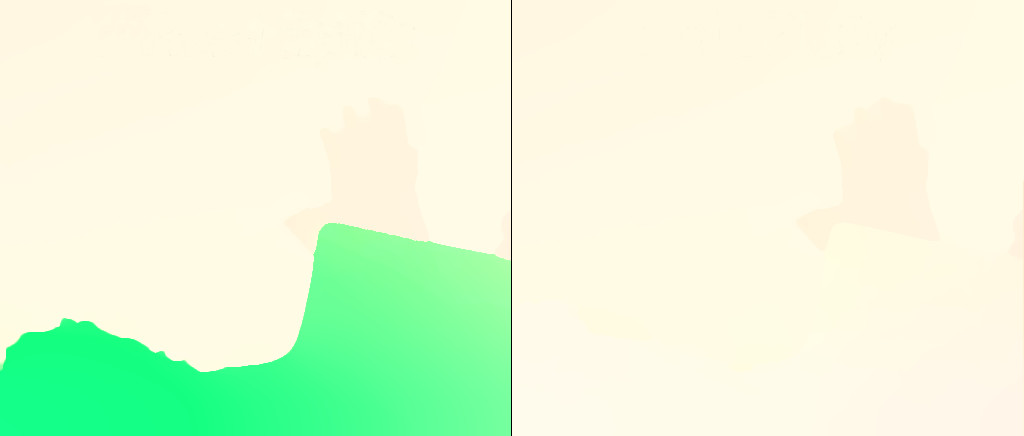}} 
 \put(3,41){ ~~~FF+Epic~~~~~~~~EpicFlow \cite{revaud:hal-01097477}} 
 \end{picture} \\
 
 \centering ~~~\begin{rotate}{90}~~~~FF+Epic\end{rotate}~~ &  \includegraphics[width=0.235\linewidth]{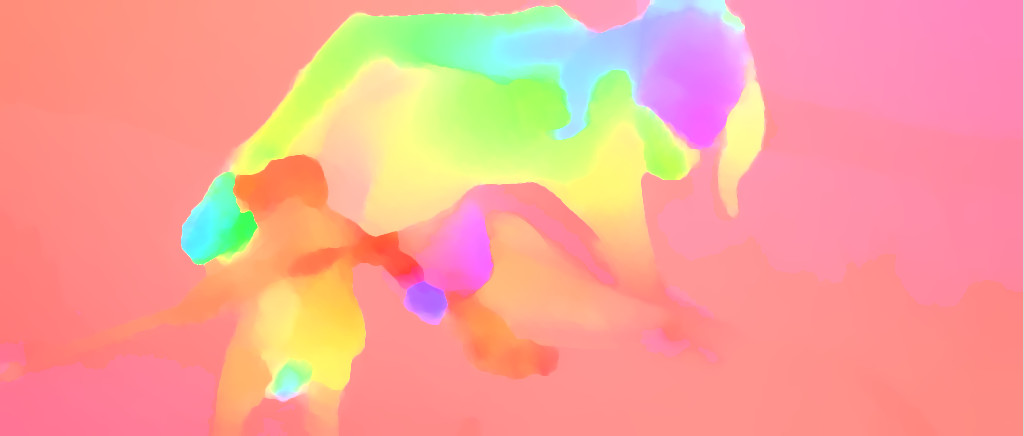} &
 \includegraphics[width=0.235\linewidth]{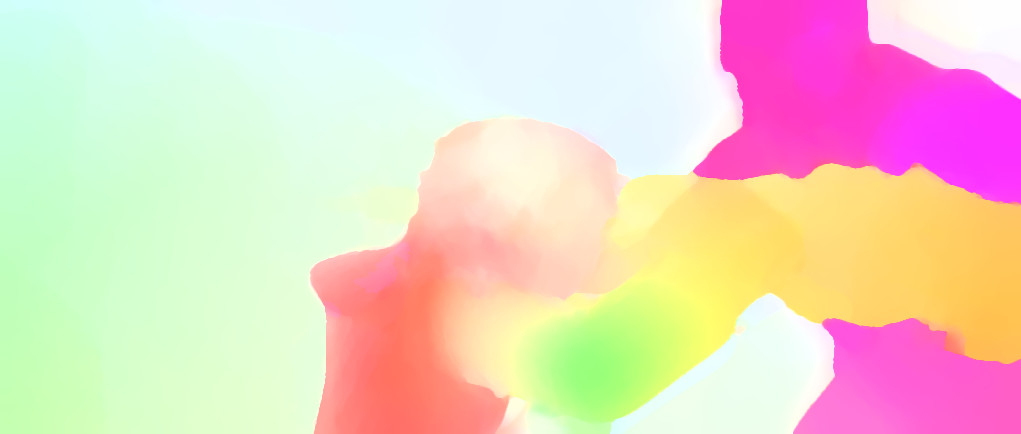} 
 & \includegraphics[height=0.100006\linewidth,trim={ {\the\alleyLeft} 0 {\the\alleyRight} 0},clip]{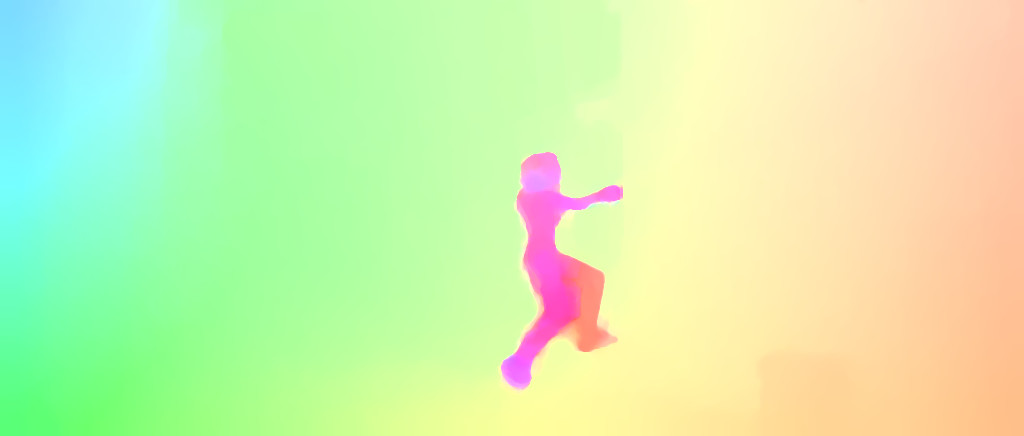}   
 & \includegraphics[height=0.100006\linewidth,trim={ {\the\alleyLeftx} 0 {\the\alleyRightx} 0},clip]{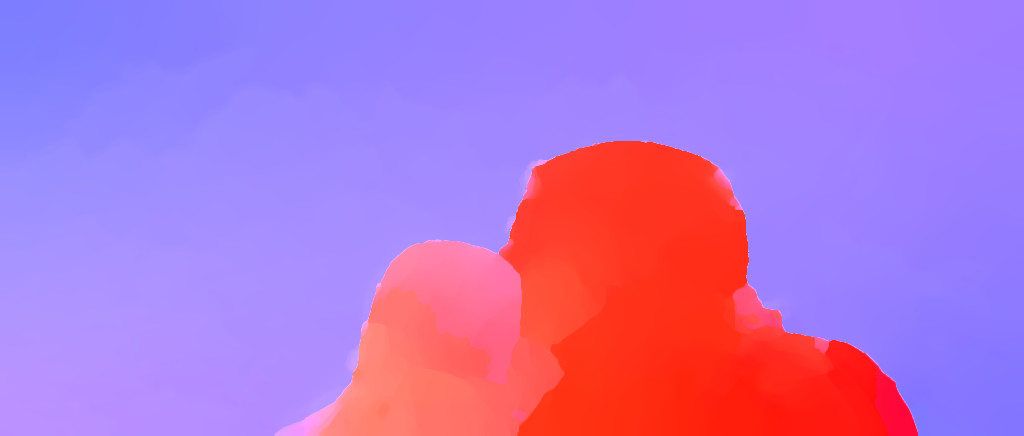} 

 & \centering ~~~~~\begin{rotate}{90}~~~~~\end{rotate}~~& 
  \begin{picture}(117,0)
  \put(0,0){\includegraphics[width=0.235\linewidth]{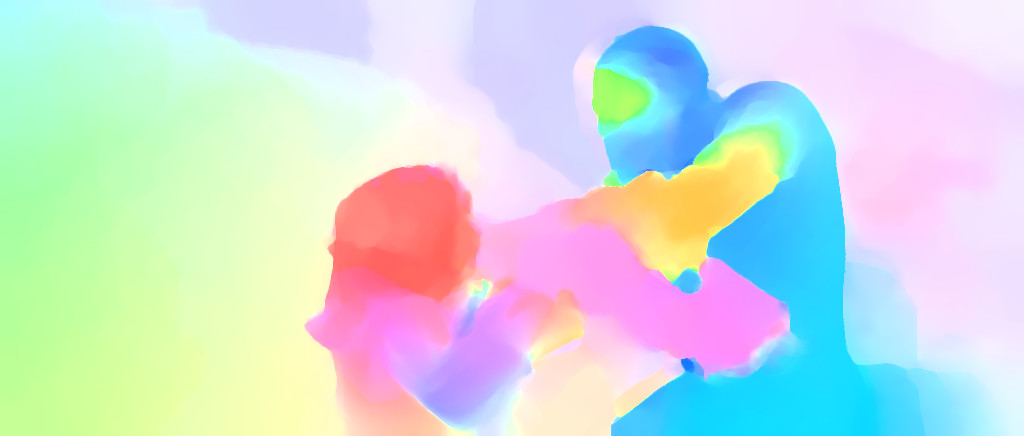}} 
  \put(3,41){ FF+Epic } 
  \end{picture} \\
 
 \centering ~~~\begin{rotate}{90}EpicFlow \cite{revaud:hal-01097477}\end{rotate}~~ &  \includegraphics[width=0.235\linewidth]{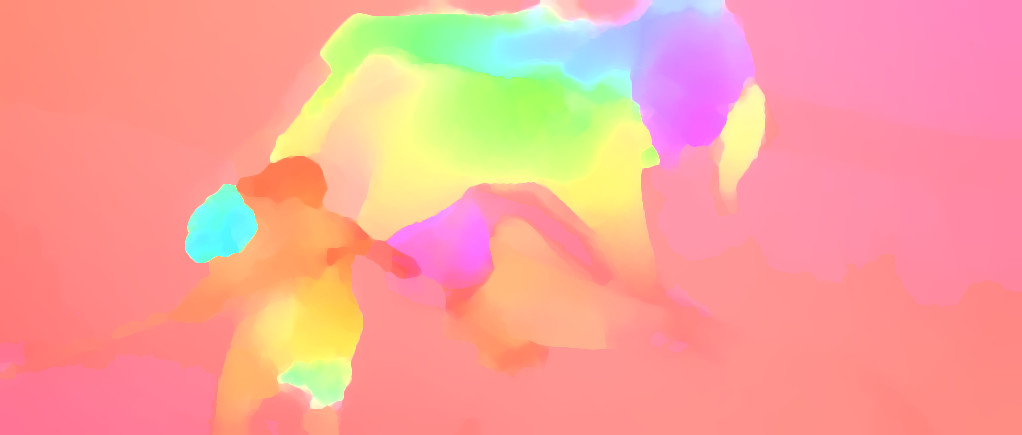} &
 \includegraphics[width=0.235\linewidth]{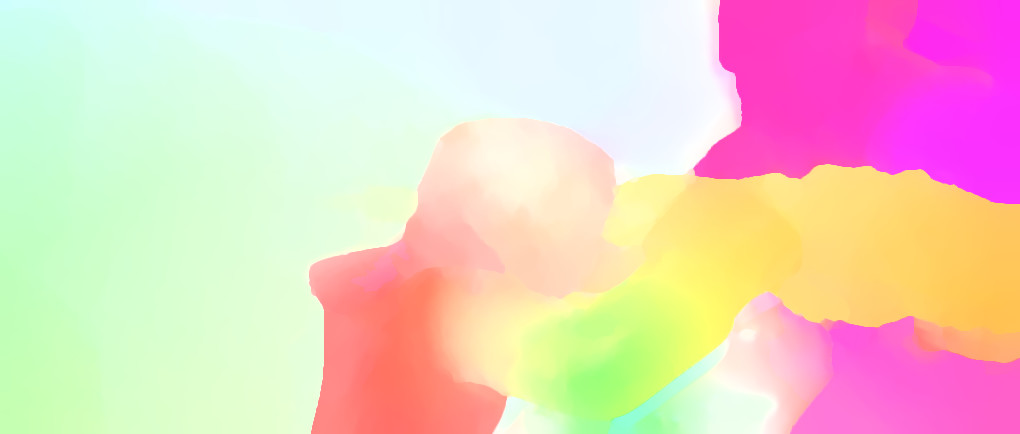} 
 & \includegraphics[height=0.100006\linewidth,trim={ {\the\alleyLeft} 0 {\the\alleyRight} 0},clip]{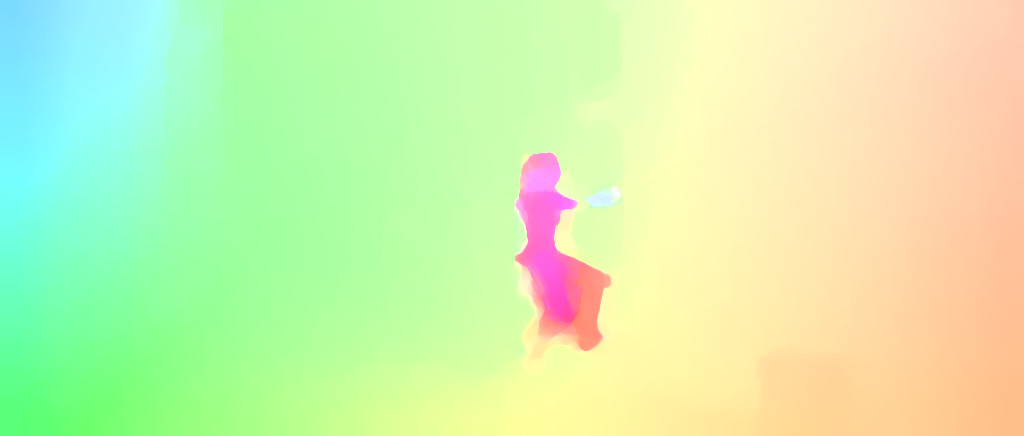} 
 & \includegraphics[height=0.100006\linewidth,trim={ {\the\alleyLeftx} 0 {\the\alleyRightx} 0},clip]{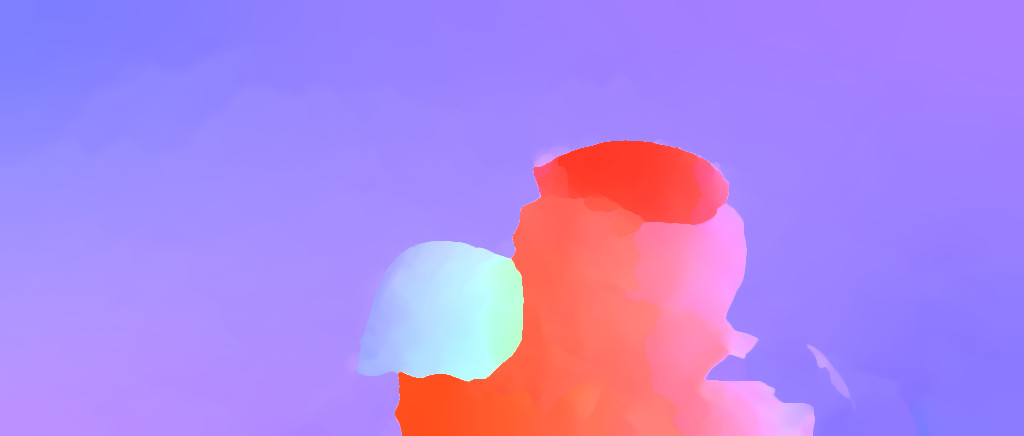}  
 
 &  \centering ~~~~~\begin{rotate}{90}~~~~\end{rotate}~~& 
  \begin{picture}(117,0)
  \put(0,0){\includegraphics[width=0.235\linewidth]{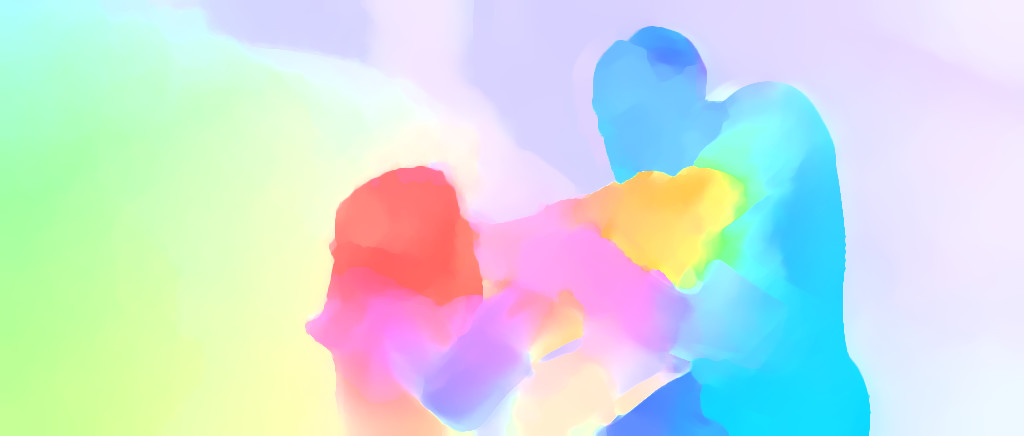}} 
  \put(3,41){ EpicFlow \cite{revaud:hal-01097477}} 
  \end{picture} \\

 \centering ~~~\begin{rotate}{90}Ground truth\end{rotate}~~ &  \includegraphics[width=0.235\linewidth]{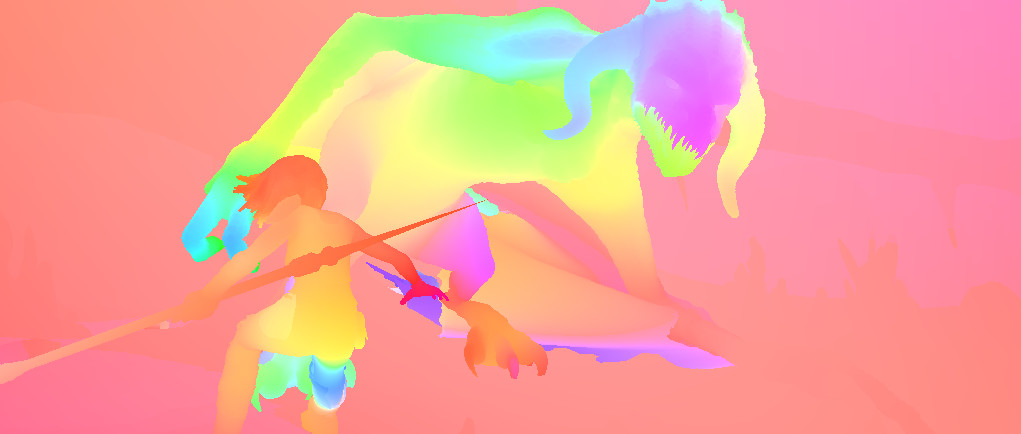} &
 \includegraphics[width=0.235\linewidth]{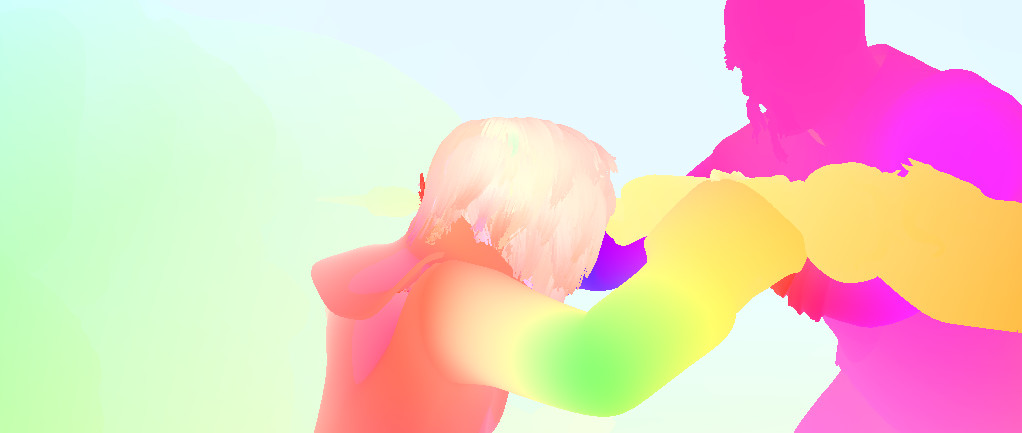} 
 & \includegraphics[height=0.100006\linewidth,trim={ {\the\alleyLeft} 0 {\the\alleyRight} 0},clip]{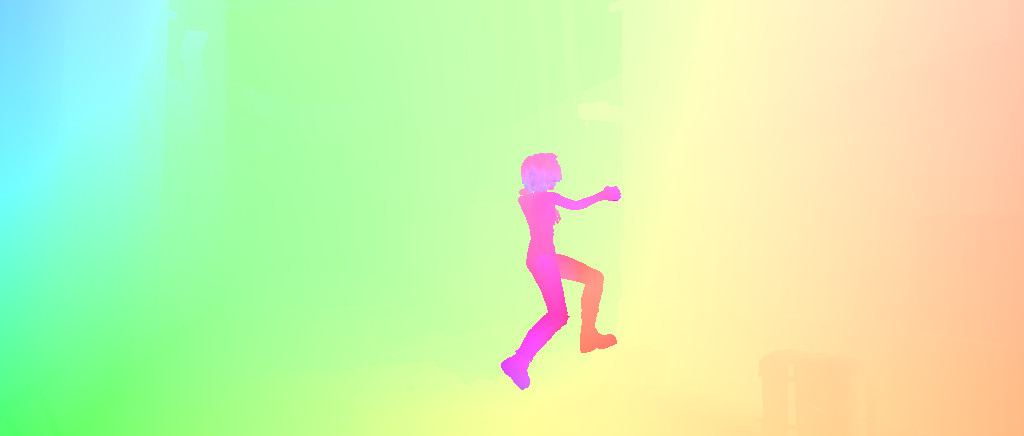}  
 & \includegraphics[height=0.100006\linewidth,trim={ {\the\alleyLeftx} 0 {\the\alleyRightx} 0},clip]{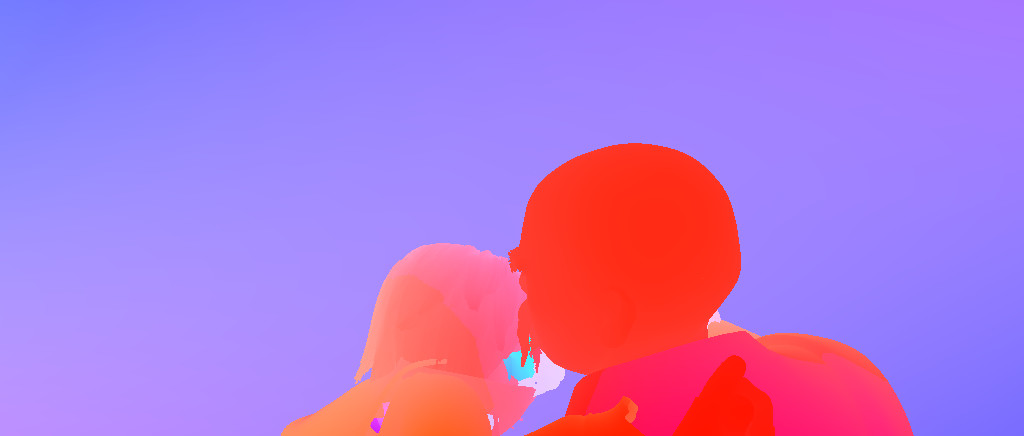}

 &  \centering ~~~~~~\begin{rotate}{90}\color{black}\textbf{a)}\color{black}~~~~~~~~~~~------~~Failure Case~~------\end{rotate}~~&
 \begin{picture}(117,0)
  \put(0,0){\includegraphics[width=0.235\linewidth]{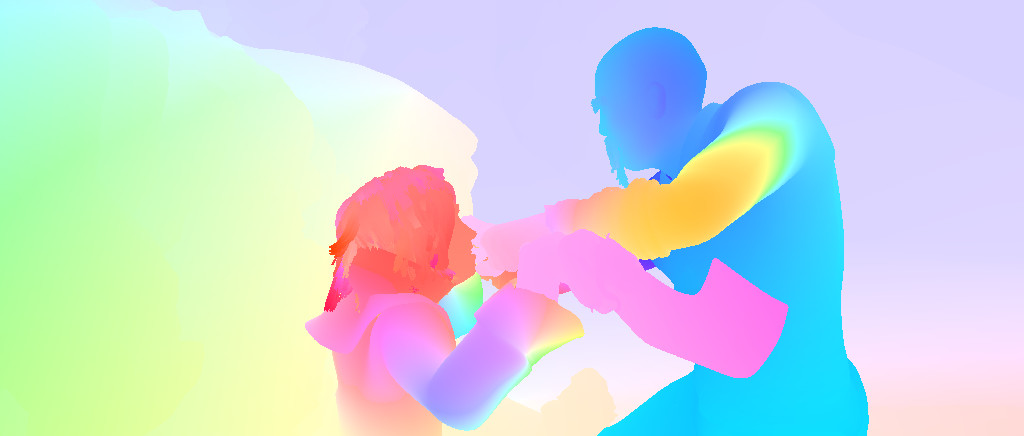}} 
  \put(3,41){ Ground truth } 
  \end{picture}\\
  
 \end{tabular}
 \vspace{0.1cm}

\captionof{figure}{The left 4 columns show example results. \textit{Images} is the average of both input images. For ANNF we use~\cite{he2012computing} 
in a fair way (see text). \textit{FF} means Flow Fields.  \textit{OM} means that the ground truth occlusion map is added (black pixels, 
it is incomplete at image boundaries). \textit{Filtered FF} is after outlier filtering (deleted pixels in black). 
\textit{FF+Epic} is EpicFlow applied on our Flow Fields. \textit{EpicFlow} is the original EpicFlow. 
Right column: a) Our approach fails in the face of the right person (outlier) and at its back (blue samples too far right).
Still our EPE is smaller due to more preserved details. b) The marked bright green flow is not considered due to too strong outlier filtering.
This makes a huge difference here. c) We show that our Flow Fields (bottom left) perform much better in presence of blur than ANNF (top left).}
  \label{visresults}
\end{table*} 

\renewcommand{\arraystretch}{1}
\renewcommand{\tabcolsep}{5pt}

\section{Evaluation \label{eva}}
We evaluate our approach on 3 optical flow datasets:
\begin{itemize}\itemsep2pt
\item MPI-Sintel~\cite{butler2012naturalistic}: It is based on an animated movie and contains many large motions up to 400 pixels per frame. 
The test set consists of two versions: \textit{clean} and \textit{final}. \textit{Clean} contains realistic illuminations and reflections. 
\textit{Final} additionally adds rendering effects like motion, defocus blurs and atmospheric effects.
\item Middlebury~\cite{baker2011database}: 
It was created for accurate optical flow estimation with relatively small displacements. Most approaches can obtain an endpoint error (EPE) in the subpixel range. 
\item KITTI~\cite{geiger2013vision}: It was created from a platform on a driving car and contains images of city streets. The motions 
can become large when the car is driving. 
\end{itemize}

In Section~\ref{exp} we perform experiments to analyze our approach and compare it to ANNF.  
In Section~\ref{res} we present our results in the public evaluation portals of the introduced datasets. 
For simplicity, we use $k=3$ and $R=1$ which we have found to perform well based on a few incoherent tests (and Table~\ref{tesin} and \ref{temid} for k), 
$l=8$ equivalent to~\cite{he2012computing} and $r=8$ and $r_2=6$ as runtime tradeoffs for the census transform. 
Only $\epsilon$ ($\pm 1$), $e$ ($\pm 1$), $s$ ($\pm 50$) and $r=r_2+1$ for SIFT flow were tuned coherently on all training frames.
$\epsilon$, $e$ and $s$ are set to 5, 4 and 50 for MPI-Sintel, to 1, 7 and 50 for Middleburry and to 1, 9 and 150 for KITTI, respectively. 
If not mentioned differently we use the census transform as data term. 
For EpicFlow applied on Flow Fields we use their standard parameters which are tuned for their Deep Matching features~\cite{weinzaepfel:hal-00873592}.
For a fair comparison we use the same parameters (tuning $\epsilon$, $e$, $s$ for ANNF does not affect our results), data term and WHTs 
in CIELab space for the ANNF approach~\cite{he2012computing} (the original approach performs even worse). 
This includes ANNF results in Section~\ref{exp} and in Figure~\ref{flowfields} and ~\ref{visresults}. 
More details regarding parameter selection and more experiments can be found in our supplementary material.

Visual results are shown in Figure~\ref{visresults}. 
EpicFlow can preserve considerably more details with our Flow Fields than with the original Deep Matching features.
Even in failure cases like in Figure~\ref{visresults} a) (right column), our approach often still achieves a smaller EPE thanks to more preserved details.  
Note that the shown failure cases also happen to the original EpicFlow. Despite more details our approach in general does not incorporate more outliers.
The occasional removal of important details like the one marked in Figure~\ref{visresults} b) remains an issue -- even for our improved outlier filtering approach. 
The marked detail is important as the flow of the very fast moving object is different on the left (brighter green). 
Still, we can in general preserve more details than the original EpicFlow.
Figure~\ref{visresults} c) shows that our approach also performs well in presence of motion and defocus blur.

\subsection{Experiments \label{exp}} 
In the introduction we claimed that our Flow Fields are better suited for optical flow estimation than ANNF and contain significantly fewer outliers.
To prove our statement quantitatively we compare our Flow Fields with different number of hierarchy levels $k$ to the state-of-the-art
ANNF approach presented in~\cite{he2012computing}. We also compare to the real NNF calculated in several days on the GPU.
The comparison is performed in Table~\ref{tesin} with 4 different measures: 

\begin{itemize}\itemsep 2.pt
\item The percentage of flows with an EPE below 3 pixels.  
\item The EPE bounded to a maximum of 10 pixels for each flow value (EPE10). 
  Outliers in correspondence fields can have arbitrary offsets, but the difficulty to remove them does not scale with their EPE. Local outliers can even 
  be more harmful since they are more likely to pass the consistency check. The EPE10 considers this. 
\item The real endpoint error (EPE) of the raw correspondence fields. It has to be taken with care (see EPE10). 
\item The EPE after outlier filtering (like in Section~\ref{out}) and utilizing EpicFlow to fill the gaps (Epic).
\end{itemize}
All 4 measures are determined in non-occluded areas only, as it is impossible to determine data based correspondences in occluded areas.
As can be seen, we can determine nearly 90\% of the pixels on the challenging MPI-Sintel training dataset with an EPE below 3 pixels, 
relying on a purely data based search strategy which considers each position in the image as possible correspondence. With weighted median filtering 
(weighted by matching error) this number can even be improved further, but the distribution is unfavorable for EpicFlow (it probably removes 
important details similar to some regularization methods).
In contrast, more hierarchy levels up to the tested $k=3$ have a positive effect on the EPE as they successfully can provide the required details. 

Bao et al.~\cite{bao2014fast} also used hierarchical matching in their approach to speed it up.
However, despite joined bilateral upsampling combined with local patch matching in a 3x3 window they found 
that the quality on Middlebury drops clearly due to hierarchical matching. As can be seen in Table~\ref{temid}
 this is not the case for our approach. As expected from the experiment in Figure~\ref{propimg} the quality even rises. Note that the Epic
result does not rise much as EpicFlow is not designed for datasets like Middlebury with EPEs in the subpixel area. 
Even with the ground truth it does not perform much better than with our approach.
Our upsampling strategy requires 11 patch comparisons while~\cite{bao2014fast} requires 9 comparisons and joined bilateral upsampling. 
However, in contrast to their upsampling strategy ours is non-local which means that we can easily correct inaccuracies and errors from a coarser level
 (the non-locality is demonstrated in Figure \ref{propimg} a)).  

\begin{figure}[t]  \vspace{-0.08cm}
\centering
  \includegraphics[width=0.881\linewidth]{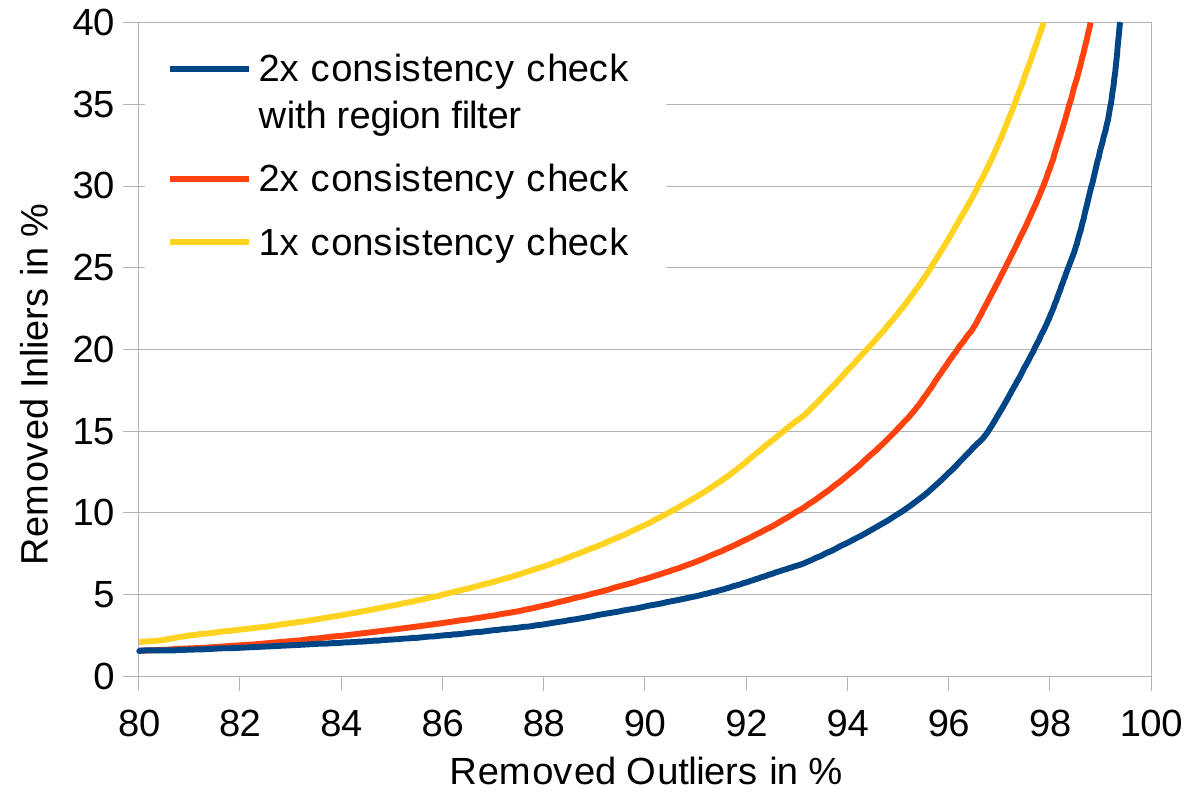}
  \vspace{-0.02cm}
   \caption{Percentage of removed outliers versus percentage of removed inliers, for an outlier threshold of 5 pixels (We vary $\epsilon$).}
\label{outliers}
\end{figure}

\paragraph{Outlier Filtering}
Figure~\ref{outliers} shows the percentage of outliers that are removed versus the percentage of inliers that are removed 
by different consistency checks on the MPI-Sintel training set.
Both the 2x consistency check as well as the region filter increase the amount of removed outliers for a fixed inlier ratio.  
We also considered using the matching error $E_d$ for outlier filtering, but there is no big gain to achieve 
(see supplementary material). 

\renewcommand{\tabcolsep}{5.0pt}
\begin{table}
\footnotesize
 \centering
 \begin{tabular}{|c|c|c|c|c|c|}
  \hline
 Method  &  $\leq 3$ pixel & EPE10 & EPE & Epic \\
 \hline
 $k=3$+median  & 92.17\% &  0.91 &  4.41 & 2.13 \\
 \hline
 $k=3$ & 89.20\% &  1.30 &  6.04 & 2.04 \\ 
 \hline
 $k=2$  & 88.79\%  & 1.36 & 8.84 & 2.08   \\ 
 \hline
 $k=1$ & 86.88\%   &  1.57 &  14.65 & 2.27 \\ 
 \hline
 $k=0$  &  79.13\% &  2.29  &   32.51  & 2.81  \\
 \hline
 ANNF \cite{he2012computing}&  68.05 \% &  3.38 & 59.11 & 3.41 \\
 \hline
 NNF &  60.20 \%  & 4.18 &  110.30 & - \tablefootnote{No backward flow calculated }\\ 
 \hline
  Original EpicFlow &\multicolumn{3}{c|}{-}  & 2.48\\ 
  \hline
 \end{tabular}
 \vspace{0.1cm}
 \caption{ Comparison of different correspondence fields on a representative subset (2x every 10th frame) on non-occluded
  regions of the MPI-Sintel training set (\textit{clean} and \textit{final}). See text for details.}
 \label{tesin}
\end{table}
  
\begin{table}
\footnotesize
 \centering
 \begin{tabular}{|c|c|c|c|c|c|}
  \hline
 Method  & $\leq 1$ pixel & EPE3 & EPE & Epic \\
 \hline
 Ground truth & 100\% & 0.0 & 0.0 &   0.214\\
 \hline
 $k=3$ & 87.08 \%  & 0.499  & 1.16  & 0.239  \\
 \hline
 $k=2$ &  86.81\% & 0.508 & 2.32 & 0.240   \\
 \hline
 $k=0$  & 81.93\% & 0.670   &  12.33    & 0.240  \\
 \hline
  Original EpicFlow &\multicolumn{3}{c|}{-}  & 0.380\\ 
  \hline
 \end{tabular}
 \vspace{0.1cm}
 \caption{Comparison of our approach with different hierarchy levels on the Middlebury training dataset to demonstrate that the quality does not suffer 
  from  hierarchical matching like in~\cite{bao2014fast}. Note that the Epic result is biased to the value in the first row.}
  \label{temid}
\end{table}

\subsection{Results \label{res}} 
\paragraph{MPI-Sintel}
Our results compared to other approaches on MPI-Sintel can be seen in Table~\ref{trsin}. We clearly outperform the original EpicFlow
as well as all other approaches. We can reduce the EPE on \textit{final} by nearly 0.5 pixels and nearly 0.4 pixels on \textit{clean}.
Most of this advance is obtained in the non-occluded area but EpicFlow also rewards our better input in the occluded areas.
On \textit{clean} we can reduce the EPE in non-occluded areas to only 1.056 pixels, which is far from the performance of most other approaches. 
On \textit{final} we can drastically reduce the error of fast motions of more than 40 pixels (s40+). 
Our approach also performs well close to occlusion boundaries (d0-10). 

\paragraph{Middlebury} \vspace{-8pt} 
On Middlebury we obtain an average rank of 38.0 (EpicFlow: 52.2) and an average EPE of 0.33 (EpicFlow: 0.39). 
Our rank is either exactly the same as EpicFlow (e.g. 69 on Army) or better  (e.g. 4 instead of 53 on Urban).
As already discussed in Section~\ref{exp} the EPE rank that can be obtained with EpicFlow on Middlebury is limited, as EpicFlow is 
not designed for such datasets. Nevertheless, we can improve the result on some datasets.  
\paragraph{KITTI} \vspace{-8pt} 
On KITTI patch based approaches seem to either perform poorly~\cite{chen2013large}, 
use scale robust features~\cite{Ranftl_ECCV2014} or special techniques like plane fitting~\cite{bao2014fast}. 
We think this is because image patches of walls and the street are undergoing strong scale changes and deformations (due to high view angle).
With the census transform our results are good for an unmodified patch based approach but not state-of-the-art (see supplementary material).
However, as our approach allows to exchange data terms as easily as parameters we use the more deformation and scale robust SIFT flow data term
to obtain the results on KITTI presented in Table~\ref{trkit}.\footnote{
Our approach with SIFT flow also outperforms EpicFlow on the
MPI-Sintel and Middlebury training sets (but less). See supplementary material.
} We use small patches with $r=3$ and $r_2=2$ as 
the benefit of SIFT to be scale and deformation robust is otherwise destroyed.
Due to the small patch sizes we use $S=12$ and $S_2=18$ for the 2. consistency check as runtime tradeoffs.
As can be seen, we just missed the best approach by 0.01\%  in \textit{$>$~3 pixel nocc}. Our approach only fails slightly in \textit{$>$3 pixel all}.
However, note that interpolation into the occluded areas is performed by EpicFlow. 
There might be better interpolation methods for the specific application of planar street scenes.
Compared to the original EpicFlow we are much better. 
Indeed, our approach is currently the only one with top performance on Sintel clean and final, as well as KITTI. 

Interesting is that although we have to use very small patches on KITTI, our hierarchical 
approach (with enlarged but blurred patches) still works very well. 
This demonstrates that the concept of hierarchical matching works even in challenging cases when matching large patches fails. 

\paragraph{Runtime} \vspace{-8pt}  Our approach including EpicFlow requires 18s for a frame in MPI-Sintel running on the 
CPU.\footnote{In detail: $ 3\times0.4s$ for kd-tree initialization, $ 2\times5s+1\times3s$ for the three Flow Fields,  $ 0.1s$ for outlier filtering and $ 3.5s$ for EpicFlow.}
By using patches with $r$ $=$ $6$ and no second consistency check we can reduce the total time to 10s with an EPE increase of only 0.13
on \textit{final} (training set) and even a decrease of 0.02 on \textit{clean} as smaller patches perform better here.
On KITTI our approach with SIFT flow needs 23 seconds per image (13 seconds without PCA). 
The best approach PPR-Flow needs 800s and the third best NLTGV-SC 16s, but on the GPU. 

\renewcommand{\tabcolsep}{1.7pt}
\begin{table}
\footnotesize
 \centering
 \begin{tabular}{|c|C{0.92 cm}|C{1.2 cm}|C{1.1 cm}|C{0.90 cm}|C{0.90 cm}|C{0.90 cm}|}
 \hline
  Method (Final) & EPE all & EPE nocc. & EPE occ. & d0-10 & s40+ \\
  \hline
  \textbf{Flow Fields} & \textbf{5.810} & \textbf{2.621} & \textbf{31.799} & \textbf{4.851}  & \textbf{33.890}  \\
  \hline
  EpicFlow \cite{revaud:hal-01097477}& \uline{6.285} & 3.060 & \uline{32.564} & 5.205  & \uline{38.021}\\
  \hline
  TF+OFM \cite{kennedy2015optical}& 6.727 &  3.388 &  33.929 &	5.544  	& 39.761 \\
  \hline
  SparseFlowFused\cite{timofte2015sparse} & 7.189 &	3.286 &	38.977  & 5.567   &  44.319  \\
  \hline
  DeepFlow \cite{weinzaepfel:hal-00873592}& 7.212 &	3.336 &	38.781 & 5.650 	 & 44.118 \\
  \hline
  NFF-Local \cite{chen2013large}& 7.249 &  \uline{2.973}  &	42.088 & \uline{4.896}    & 44.866\\
  \hline
  \hline
  Method (Clean) & EPE all & EPE nocc. & EPE occ. & d0-10 & s40+ \\
  \hline
  \textbf{Flow Fields} & \textbf{3.748} & \textbf{1.056} & \textbf{25.700} & \uline{2.784}  & \textbf{23.602}  \\
  \hline
  EpicFlow \cite{revaud:hal-01097477}& \uline{4.115} & \uline{1.360} & 26.595 & 3.660 & \uline{25.859}\\
  \hline
  PH-Flow \cite{PhFlow} &  4.388  & 1.714 & \uline{26.202} & 3.612  & 27.997\\
  \hline
  NNF-Local\cite{chen2013large}& 5.386 & 1.397 & 37.896 & \textbf{2.722} & 36.342 \\
  \hline
 \end{tabular}
 \vspace{0.1cm}
 \caption{Results on MPI-Sintel. Bold results are the best, underlined the 2. best.
  (n)occ = (non) occluded. d0-10 = 0 - 10 pixels from occlusion boundary. s40+ = motions of more than 40 pixels.}
 \label{trsin}
\end{table}

\begin{table}
\footnotesize
 \centering
 \begin{tabular}{|c|C{0.75 cm}|C{1.2 cm}|C{1.2 cm}|C{0.9 cm}|C{0.9 cm}|}
  \hline
  Method  & Rank & $>$3 pixel nocc.&  $>$3 pixel all& EPE nocc. & EPE all  \\
  \hline 
  PH-Flow   \cite{PhFlow}& 1& \textbf{5.76 \%} & \textbf{10.57 \%} & \textbf{1.3 px} & \textbf{2.9 px} \\ 
  \hline
  \textbf{Flow Fields} & 2  &  \uline{5.77 \%} & 14.01 \% & \uline{1.4 px} &  \uline{3.5 px}  \\ 
  \hline
  NLTGV-SC   \cite{Ranftl_ECCV2014}& 3 &  5.93 \% & \uline{11.96 \%} & 1.6 px & 3.8 px  \\
  \hline
  DDS-DF  \cite{wei2014datadriven}& 4 & 6.03 \% &13.08 \% &	1.6 px 	& 4.2 px \\  
  \hline
   TGV2ADCSIFT  \cite{brauxzin2013iccv}& 5& 6.20 \% &	15.15 \% & 1.5 px & 4.5 px \\ 
  \hline
  EpicFlow \cite{revaud:hal-01097477} & 13 & 7.88 \% & 17.08 \% &  1.5 px & 3.8 px\\
   \hline
 \end{tabular}
 \vspace{0.1cm}
 \caption{Results on KITTI test set. The table rank is the original rank excluding non optical flow methods. 
  nocc. = Non-occluded.}
 \label{trkit}
\end{table}
\renewcommand{\tabcolsep}{5pt}

\section{Conclusion and Future Work \label{con}}
In this paper we presented a novel correspondence field approach for optical flow estimation. 
We showed that our Flow Fields are clearly superior to ANNF and better suited than state-of-the-art descriptor matching approaches, regarding optical flow estimation.
We also presented advanced outlier filtering and demonstrated that we can obtain promising optical flow results, utilizing a state-of-the-art optical flow algorithm 
like EpicFlow. With our results, we hope to inspire the research of dense correspondence field estimation for optical flow. 
So far, sparse descriptor matching techniques are much more popular as too little effort was spent in improving dense techniques.

In future work, more advanced data terms can be tested. Thanks to intensive research mainly in stereo estimation
there are nowadays e.g. many improvements for the census transform~\cite{demetz2013complete, Ranftl_ECCV2014, luan2012illumination, xiong2010color}. 
These can probably be used to further improve our approach. Promising is also to estimate patch deformations by random search~\cite{hacohen2011non}. It is known that 
this works well for patch normals in 3D reconstruction~\cite{bailer2012scale}.
 \vspace{-1pt}
\subsection*{Acknowledgements} \vspace{-1pt} This work was partially funded by the BMBF project DYNAMICS (01IW15003) and the EU 7th Framework Programme project AlterEgo (600610). 
{\small
\bibliographystyle{ieee}
\bibliography{paper}

\begin{thebibliography}{10}\itemsep=-1pt

\bibitem{bailer2012scale}
C.~Bailer, M.~Finckh, and H.~P. Lensch.
\newblock Scale robust multi view stereo.
\newblock In {\em ECCV}, pages 398--411. Springer, 2012.

\bibitem{baker2011database}
S.~Baker, D.~Scharstein, J.~Lewis, S.~Roth, M.~J. Black, and R.~Szeliski.
\newblock A database and evaluation methodology for optical flow.
\newblock {\em IJCV}, 92(1):1--31, 2011.

\bibitem{bao2014fast}
L.~Bao, Q.~Yang, and H.~Jin.
\newblock Fast edge-preserving patchmatch for large displacement optical flow.
\newblock In {\em CVPR}, pages 3534--3541. IEEE, 2014.

\bibitem{barnes2010generalized}
C.~Barnes, E.~Shechtman, D.~B. Goldman, and A.~Finkelstein.
\newblock The generalized patchmatch correspondence algorithm.
\newblock In {\em ECCV}, pages 29--43. Springer, 2010.

\bibitem{brauxzin2013iccv}
J.~Braux-Zin, R.~Dupont, and A.~Bartoli.
\newblock A general dense image matching framework combining direct and
  feature-based costs.
\newblock In {\em ICCV}. IEEE, 2013.

\bibitem{brox2004high}
T.~Brox, A.~Bruhn, N.~Papenberg, and J.~Weickert.
\newblock High accuracy optical flow estimation based on a theory for warping.
\newblock In {\em ECCV}, pages 25--36. Springer, 2004.

\bibitem{brox2011large}
T.~Brox and J.~Malik.
\newblock Large displacement optical flow: descriptor matching in variational
  motion estimation.
\newblock {\em PAMI}, 33(3):500--513, 2011.

\bibitem{butler2012naturalistic}
D.~J. Butler, J.~Wulff, G.~B. Stanley, and M.~J. Black.
\newblock A naturalistic open source movie for optical flow evaluation.
\newblock In {\em ECCV}, pages 611--625. Springer, 2012.

\bibitem{chen2013large}
Z.~Chen, H.~Jin, Z.~Lin, S.~Cohen, and Y.~Wu.
\newblock Large displacement optical flow with nearest neighbor fields.
\newblock In {\em CVPR}, pages 2443--2450. IEEE, 2013.

\bibitem{demetz2013complete}
O.~Demetz, D.~Hafner, and J.~Weickert.
\newblock The complete rank transform: A tool for accurate and morphologically
  invariant matching of structures.
\newblock In {\em BMVC}, 2013.

\bibitem{duchon1979lanczos}
C.~E. Duchon.
\newblock Lanczos filtering in one and two dimensions.
\newblock {\em Journal of Applied Meteorology}, 18(8):1016--1022, 1979.

\bibitem{furukawa2010accurate}
Y.~Furukawa and J.~Ponce.
\newblock Accurate, dense, and robust multiview stereopsis.
\newblock {\em PAMI}, 32(8):1362--1376, 2010.

\bibitem{geiger2013vision}
A.~Geiger, P.~Lenz, C.~Stiller, and R.~Urtasun.
\newblock Vision meets robotics: The kitti dataset.
\newblock {\em The International Journal of Robotics Research}, 2013.

\bibitem{goesele2007multi}
M.~Goesele, N.~Snavely, B.~Curless, H.~Hoppe, and S.~M. Seitz.
\newblock Multi-view stereo for community photo collections.
\newblock In {\em ICCV}, pages 1--8. IEEE, 2007.

\bibitem{hacohen2011non}
Y.~HaCohen, E.~Shechtman, D.~B. Goldman, and D.~Lischinski.
\newblock Non-rigid dense correspondence with applications for image
  enhancement.
\newblock {\em ACM transactions on graphics (TOG)}, 30(4):70, 2011.

\bibitem{he2012computing}
K.~He and J.~Sun.
\newblock Computing nearest-neighbor fields via propagation-assisted kd-trees.
\newblock In {\em CVPR}, pages 111--118. IEEE, 2012.

\bibitem{hel2005real}
Y.~Hel-Or and H.~Hel-Or.
\newblock Real-time pattern matching using projection kernels.
\newblock {\em PAMI}, 27(9):1430--1445, 2005.

\bibitem{horn1981determining}
B.~K. Horn and B.~G. Schunck.
\newblock Determining optical flow.
\newblock In {\em 1981 Technical Symposium East}, pages 319--331. International
  Society for Optics and Photonics, 1981.

\bibitem{jith2014optical}
O.~Jith, S.~A. Ramakanth, and R.~V. Babu.
\newblock Optical flow estimation using approximate nearest neighbor field
  fusion.
\newblock In {\em Acoustics, Speech and Signal Processing (ICASSP)}, pages
  673--674. IEEE, 2014.

\bibitem{kennedy2015optical}
R.~Kennedy and C.~J. Taylor.
\newblock Optical flow with geometric occlusion estimation and fusion of
  multiple frames.
\newblock In {\em Energy Minimization Methods in Computer Vision and Pattern
  Recognition}, pages 364--377. Springer, 2015.

\bibitem{korman2011coherency}
S.~Korman and S.~Avidan.
\newblock Coherency sensitive hashing.
\newblock In {\em ICCV}, pages 1607--1614. IEEE, 2011.

\bibitem{liu2011sift}
C.~Liu, J.~Yuen, and A.~Torralba.
\newblock Sift flow: Dense correspondence across scenes and its applications.
\newblock {\em PAMI}, 33(5):978--994, 2011.

\bibitem{lu2013patch}
J.~Lu, H.~Yang, D.~Min, and M.~N. Do.
\newblock Patch match filter: Efficient edge-aware filtering meets randomized
  search for fast correspondence field estimation.
\newblock In {\em CVPR}, pages 1854--1861. IEEE, 2013.

\bibitem{luan2012illumination}
X.~Luan, F.~Yu, H.~Zhou, X.~Li, D.~Song, and B.~Wu.
\newblock Illumination-robust area-based stereo matching with improved census
  transform.
\newblock In {\em Measurement, Information and Control (MIC)}, volume~1, pages
  194--197. IEEE, 2012.

\bibitem{Ranftl_ECCV2014}
R.~Ranftl, K.~Bredies, and T.~Pock.
\newblock Non-local total generalized variation for optical flow estimation.
\newblock In {\em ICCV}, 2014.

\bibitem{revaud:hal-01097477}
J.~Revaud, P.~Weinzaepfel, Z.~Harchaoui, and C.~Schmid.
\newblock {EpicFlow: Edge-Preserving Interpolation of Correspondences for
  Optical Flow}.
\newblock In {\em CVPR}, 2015.

\bibitem{sun2010secrets}
D.~Sun, S.~Roth, and M.~J. Black.
\newblock Secrets of optical flow estimation and their principles.
\newblock In {\em CVPR}, pages 2432--2439. IEEE, 2010.

\bibitem{timofte2015sparse}
R.~Timofte and L.~Van~Gool.
\newblock Sparse flow: Sparse matching for small to large displacement optical
  flow.
\newblock In {\em Applications of Computer Vision (WACV)}, pages 1100--1106.
  IEEE, 2015.

\bibitem{vogel2013evaluation}
C.~Vogel, S.~Roth, and K.~Schindler.
\newblock An evaluation of data costs for optical flow.
\newblock In {\em Pattern Recognition (GCPR)}, pages 343--353. Springer, 2013.

\bibitem{Wedel2009}
A.~Wedel, D.~Cremers, T.~Pock, and H.~Bischof.
\newblock Structure-and motion-adaptive regularization for high accuracy optic
  flow.
\newblock In {\em ICCV}, pages 1663--1668. IEEE, 2009.

\bibitem{wei2014datadriven}
D.~Wei, C.~Liu, and W.~Freeman.
\newblock A data-driven regularization model for stereo and flow.
\newblock In {\em 3DTV-Conference}. IEEE, 2014.

\bibitem{weinzaepfel:hal-00873592}
P.~Weinzaepfel, J.~Revaud, Z.~Harchaoui, and C.~Schmid.
\newblock {DeepFlow: Large displacement optical flow with deep matching}.
\newblock In {\em ICCV}, 2013.

\bibitem{xiong2010color}
G.~Xiong, X.~Li, J.~Gong, H.~Chen, and D.-J. Lee.
\newblock Color rank and census transforms using perceptual color contrast.
\newblock In {\em Control Automation Robotics \& Vision (ICARCV)}, pages
  1225--1230. IEEE, 2010.

\bibitem{xu2012motion}
L.~Xu, J.~Jia, and Y.~Matsushita.
\newblock Motion detail preserving optical flow estimation.
\newblock {\em PAMI}, 34(9):1744--1757, 2012.

\bibitem{PhFlow}
J.~Yang and H.~Li.
\newblock Dense, accurate optical flow estimation with piecewise parametric
  model.
\newblock In {\em CVPR}, pages 1019--1027, 2015.

\bibitem{zabih1994non}
R.~Zabih and J.~Woodfill.
\newblock Non-parametric local transforms for computing visual correspondence.
\newblock In {\em ECCV}, pages 151--158. Springer, 1994.

\end{thebibliography}
}
\end{document}